\documentclass[letterpaper,twocolumn,10pt]{article}
\usepackage{usenix}

\usepackage{mdframed,lipsum}
\usepackage{amsmath,amsfonts}
\usepackage{graphicx}
\usepackage{xcolor}
\usepackage{textcomp}
\usepackage{bm}
\usepackage{balance}  
\usepackage{float}
\usepackage{xcolor}
\usepackage{color, colortbl}
\usepackage{diagbox}
\usepackage{amsmath}
\usepackage{url}
\usepackage{multirow}
\usepackage[ruled,linesnumbered, vlined]{algorithm2e}
\usepackage{subcaption}
\usepackage{mathtools}
\usepackage{amsfonts}
\usepackage{threeparttable}
\usepackage{booktabs}
\usepackage{enumitem}
\usepackage{subfiles}

\newcommand{\TargetM}{\Phi}
\newcommand{\TargetG}{G}
\newcommand{\SyntheticG}{\hat{G}}
\newcommand{\ShadowG}{\TargetG^{S}}
\newcommand{\ShadowM}{\TargetM^{S}}

\newcommand{\Ttrain}{$\Phi^{\text{train}}$}
\newcommand{\Ttest}{$\Phi^{\text{test}}$}
\newcommand{\Atrain}{$\mathcal{A}^{\text{train}}$}

\newcommand{\ExternalK}{\mathbb{K}}
\newcommand{\Property}{\mathbb{P}}

\def\D{{\bf D}}

\def\G{{\bf G}}

\def\P{{\bf P}}

\def\x{{\bf x}}

\def\0{{\bf 0}}
\def\1{{\bf 1}}

\usepackage{colortbl}
\definecolor{asparagus}{rgb}{0.53, 0.66, 0.42}
\definecolor{carrotorange}{rgb}{0.93, 0.57, 0.13}
\definecolor{lightthulianpink}{rgb}{0.9, 0.56, 0.67}

\newenvironment{ditemize}{%
\begin{list}{{\bf $\bullet$}}{%
\setlength{\itemsep}{0pt}\setlength{\rightmargin}{0pt}%
\setlength{\leftmargin}{1.2em}\setlength{\parsep}{0pt}}}{
\end{list}}

\begin{document}

\title{Inference Attacks Against Graph Generative Diffusion Models}
\author{
{\rm Xiuling Wang}\\
Hong Kong Baptist University\\
xiulingwang@hkbu.edu.hk
\and
{\rm Xin Huang}\\
Hong Kong Baptist University\\
xinhuang@comp.hkbu.edu.hk
\and
{\rm Guibo Luo}\\
Peking University\\
luogb@pku.edu.cn
\and
{\rm Jianliang Xu\thanks{Jianliang Xu is the corresponding author.}}\\
Hong Kong Baptist University\\
xujl@comp.hkbu.edu.hk
} 

\maketitle

\begin{abstract}
Graph generative diffusion models have recently emerged as a powerful paradigm for generating complex graph structures, effectively capturing intricate dependencies and relationships 
within graph data. However, the privacy risks associated with these models remain largely unexplored. In this paper, we investigate information leakage in such models through 
three types of black-box inference attacks. First, we design a graph reconstruction attack, which can reconstruct graphs structurally similar to those training graphs from the generated graphs. Second, we propose a property inference attack to infer the properties of the training graphs, such as the average graph density and the distribution of densities, from the generated graphs. Third, we develop two membership inference attacks to determine whether a given graph is present in the training set. Extensive experiments on three different types of graph generative diffusion models and six real-world graphs demonstrate the effectiveness of these attacks, significantly outperforming the baseline approaches.
Finally, we propose two defense mechanisms that mitigate these inference attacks and achieve a better trade-off between defense strength and target model utility than existing methods. Our code is available at https://zenodo.org/records/17946102.

\end{abstract}




\vspace{-0.05in}
\section{Introduction}
\vspace{-0.05in}

Many real-world systems, such as social networks, biological networks, and information networks, can be represented as graphs. Graph learning is crucial for analyzing these systems because of its ability to model and analyze complex relationships and interactions within the data.

Graph generation, a critical task in graph learning, focuses on creating graphs that accurately reflect the underlying structure of graph data. 
These models have diverse applications, such as recommender systems \cite{jiang2024diffkg, wang2023diffusion}, social network analysis \cite{liu2024score, li2024recdiff}, molecular research \cite{huang2023conditional, zhang2023survey}, and drug discovery \cite{torge2023diffhopp, wu2025graph}. 
Diffusion models have recently gained significant attention as a prominent class of generative models, which work through two interconnected processes. A forward process gradually adds noise to data until it conforms to a predefined prior distribution (e.g., Gaussian). A corresponding reverse process uses a trained neural network to
progressively denoise the data, effectively reversing the forward process and reconstructing the original data distribution. Given the success of diffusion models in image generation \cite{ho2020denoising}, there has been growing interest in applying these techniques to graph generation~\cite{fan2023generative,zhang2023survey, yang2023diffusion, cao2024survey, liu2023generative}. 

While graph generative diffusion models (GGDMs) are capable of producing various graphs, they often require large training datasets for robust generation. However, these datasets may contain sensitive or confidential information. For example, social graphs can expose private relationships, as seen in cases such as Facebook’s famous Cambridge Analytica scandal; graphs depicting protein-protein interactions or gene regulatory networks may include proprietary research data, any leakage of this information could compromise competitive advantages in biotechnology research; graphs in healthcare datasets can model relationships between patients, treatments, and healthcare providers, where the disclosure of such information could compromise patient confidentiality and violate regulations such as HIPAA \cite{moore2019review}. Therefore, this paper investigates the critical question: \emph{how much information about the training data can be inferred from GGDMs?}

{\bf Challenges}. Recent research has uncovered several types of attacks that can infer sensitive information from the training data of graph learning models, including membership inference \cite{he2021node, he2021stealing, wu2022linkteller, duddu2020quantifying, shen2022finding, wang2023link, wang2024gcl, wang2024subgraph}, attribute inference \cite{duddu2020quantifying, gong2018attribute, zhang2021graph}, property inference \cite{wang2022group, zhang2021leakage, zhang2021inference, suri2021formalizing}, and graph reconstruction attacks \cite{zhang2021graphmi, shen2022finding, zhou2023strengthening}. Most of these attacks are based on model’s final output (probability vectors) for node/graph classification \cite{he2021node, he2021stealing, wang2024subgraph, gong2018attribute, wang2022group, zhang2021leakage, suri2021formalizing}, node/graph embeddings \cite{zhang2021graph, wang2024gcl, duddu2020quantifying, shen2022finding, wang2024gcl, wang2023link, duddu2020quantifying, zhang2021inference, zhou2023strengthening}, or other features such as model gradients \cite{wu2022linkteller, zhang2021graphmi}. These approaches are not well-suited for GGDMs for two reasons:
{\em First}, attack features derived from probability vectors or node/graph embeddings are unavailable in GGDMs that generate graphs, making existing attacks infeasible in black-box setting.
{\em Second}, the above attacks rely on a one-to-one mapping between an input graph and an output to form a (feature, label) pair, where label denotes membership or a specific property. GGDMs lack this mapping as the graphs are generated from a set of training samples, making ground-truth labels unavailable and existing attacks inapplicable.
These differences highlight the unique challenge of attacking GGDMs: constructing meaningful features from generated graphs without one-to-one mappings.

On the other hand, recent preliminary studies have begun to investigate the privacy vulnerabilities of image and text-to-image generative diffusion models under various privacy inference attacks. These include membership inference attacks \cite{carlini2023extracting, duan2023diffusion, pang2023white}, which aim to determine whether a particular data sample is in the training set; data reconstruction attacks, which attempt to recover the training images \cite{carlini2023extracting}; and property inference attacks \cite{luo2024exploring}, which seek to infer the portion of training images with a specific property. However, all of the above works focus only on image or text-to-image diffusion models and cannot be directly applied to GGDMs due to the unique structural characteristics of graph data. Specifically, graph data exhibit variable sizes and topologies, permutation-invariant, and encode sensitive information through higher-order structural patterns rather than fixed spatial features.

{\bf Our Contributions}. We initiate a systematic investigation into the privacy risks of GGDMs by exploring three types of inference attacks. First, we explore {\em graph reconstruction attack} (GRA), which attempts to infer graph structures within model's training set. For example, if the target graph is from a medical database, a reconstructed graph could enable an adversary to gain knowledge of sensitive relationships between patients, health records, and treatments.
Second, we introduce {\em property inference attack} (PIA), which leverages generated graphs to infer statistical properties of the training graphs, such as average graph density or the proportion of graphs within specific density ranges. Revealing these properties may violate the intellectual property (IP) of the data owner, particularly in domains like molecular or protein graphs from biomedical companies.
Third, we investigate {\em membership inference attack} (MIA), which aims to determine whether a given graph is present in GGDM’s training set. For example, in a collaborative training setting where each data owner possesses graphs from different biological companies, with each graph representing a proprietary product, an adversary may attempt to infer whether a specific product is included in another owner's training set, thereby compromising IP.

To the best of our knowledge, this is the first
work to explore the privacy leakage of GGDMs. Overall, we make the following contributions in this paper:
\begin{itemize}[leftmargin=*]
\item We develop a novel graph reconstruction attack by aligning each generated graph with its closest counterpart in the generated graph set to identify overlapping edges. These overlapping edges are then considered part of a graph from the target model's training set. We evaluate the effectiveness of our proposed attack on six real-world graph datasets and three state-of-the-art GGDMs. The results show that the attack achieves an F1 score of up to 0.99, with up to 36\% of the original training graphs being exactly recovered.
\vspace{-0.1in}
\item We launch our property inference attack using a simple yet efficient method that directly calculates property values from the generated graphs. Extensive experiments show that the proposed attack can accurately infer the statistical properties of the training graphs. For example, on the IMDB-MULTI dataset, the difference between the actual and inferred average graph degrees can be as small as 0.005.
\vspace{-0.25in}
\item We design our membership inference attacks by employing the shadow-model-training technique, where we train MLP attack models based on two factors: (1) different similarity levels between the generated graphs and their corresponding training graphs for member and non-member graphs; and (2) different similarity levels within the generated graphs from member and non-member graphs. Experimental results
show that our attacks can achieve an AUC of up to 0.999 when shadow and target graphs are drawn from the same dataset, and 0.895 when drawn from different datasets.  
\vspace{-0.25in}
\item To mitigate the inference attacks, we propose two defense mechanisms that introduce noise into either the training or generated graphs of the target GGDM, thereby altering the model's outputs. Notably, we limit perturbations to the least significant edges and non-edges (by flipping them), to minimize the impact on the target model's utility. Empirical evaluations show that our approach achieves defense effectiveness comparable to two baseline methods, rendering the attack ineffective. Furthermore, our method achieves a better trade-off between defense strength and target model utility compared to the two existing defense methods.
\end{itemize}
\vspace{-0.1in}

\vspace{-0.05in}
\section{Related Work}
\vspace{-0.05in}
\label{sc:relat}

\begin{table}[t!]
     \centering
     \scalebox{0.78}
     {
     \begin{tabular}{|c|c|c|c|c|c|c|}\hline
     \multirow{2}{*}{}&\multicolumn{3}{c|}{\bf Attack type} &\multicolumn{2}{c|}{\bf Attack setting} &\multirow{2}{*}{\bf Domain}  \\\cline{2-6}
     &{\bf DR} &{\bf PIA} & {\bf MIA} &{\bf B-box}&{\bf W-box}&\\\hline
     \cite{carlini2023extracting}&\checkmark&&\checkmark&\checkmark&\checkmark&\\\cline{1-6}
    \cite{luo2024exploring}&&\checkmark&&\checkmark&&Image or  \\\cline{1-6}
    \cite{wu2022membership, zhang2024generated, pang2023black, li2024towards} &&&\checkmark&\checkmark&&text-to-\\\cline{1-6}
    \cite{duan2023diffusion, tang2023membership, kong2023efficient, zhai2025membership, hu2023membership} \multirow{2}{*}{}&\multirow{2}{*}{}&\multirow{2}{*}{}&\multirow{2}{*}{\checkmark}&\multirow{2}{*}{}&\multirow{2}{*}{\checkmark}&image\\
    \cite{carlini2023extracting, matsumoto2023membership, dubinski2024towards, pang2023white}&&&&&&\\\hline
    Ours&\checkmark&\checkmark&\checkmark&\checkmark&&Graph\\\hline
     \end{tabular}
     }
      \vspace{-0.05in}
     \caption{\label{tab:comp-literature} Comparison between the existing works on inference attacks against generative diffusion models. "DR", "PIA", and "MIA" refer to data reconstruction, property inference, and membership inference attacks, respectively. "B-box" and "W-box" represent black-box and white-box, respectively.}
     \vspace{-0.2in}
\end{table} 
{\bf Generative diffusion models for graphs.}
Generative diffusion models have recently gained significant attention as a powerful paradigm for deep graph generation, aiming to learn the underlying graph distribution and synthesize novel graphs.
The existing graph generative diffusion models (GGDMs) can be broadly categorized into three classes \cite{zhang2023survey, fan2023generative}: (1) {\em Score-based Generative Models (SGM)} \cite{niu2020permutation, chen2022nvdiff} that employ a score function to represent the probability distribution of the data; 
(2) {\em Denoising Diffusion Probabilistic Models (DDPMs)} \cite{haefeli2022diffusion} that add discrete Gaussian noise to the graph with Markov transition kernels \cite{austin2021structured} and then train a neural network to predict the added noise to recover the original graph, \cite{vignac2022digress} adds discrete noise instead of continuous Gaussian, and 
(3) {\em Stochastic Differential Equation-based Models (SDEs)} \cite{huang2022graphgdp, jo2022score, luo2022fast} that characterize the development of a system over time under the influence of random noise. 
SGM and DDPM leverage the score-matching idea and non-equilibrium thermodynamics, respectively, to learn different reverse functions of the diffusion process, while SDE generalizes the discrete diffusion steps into continuous scenarios and further models the diffusion process with stochastic differential equations \cite{fan2023generative}.
We refer the readers to comprehensive surveys on GGDMs \cite{fan2023generative,zhang2023survey}.

{\bf Inference attacks against generative diffusion models.} Few studies \cite{carlini2023extracting, van2021memorization, somepalli2023understanding, luo2024exploring, truong2024attacks} have investigated the privacy vulnerabilities of generative diffusion models against privacy inference attacks. Van et al. \cite{van2021memorization} focus on probabilistic deep generative models such as variational autoencoders and formulate the concept of "memorization score" by measuring the impact of removing an observation on a given model. Carlini et al. \cite{carlini2023extracting} consider image diffusion models and design three attacks: data extraction, data reconstruction, and membership inference attack. Somepalli et al. \cite{somepalli2023understanding} focus on text-to-image diffusion models and analyze the data duplication problem in these models. Luo et al. \cite{luo2024exploring} introduce a black-box property inference attack that aims to infer the distribution of specific properties from generated images. However, none of these previous works have examined the privacy leakage in GGDMs. 
Several works focus on membership inference attacks against diffusion models \cite{duan2023diffusion, pang2023white, dubinski2024towards, matsumoto2023membership, hu2023membership, zhai2025membership, kong2023efficient, tang2023membership, li2024towards, zhang2024generated, wu2022membership, pang2023black, li2024towards}. Specifically, \cite{wu2022membership, zhang2024generated, pang2023black, li2024towards} describe black-box attacks by analyzing the generated images; 
\cite{duan2023diffusion, tang2023membership, kong2023efficient, zhai2025membership, hu2023membership} and \cite{carlini2023extracting, matsumoto2023membership, dubinski2024towards} design white-box attacks that rely on posterior estimation errors and model losses, respectively; 
and Pang et al. \cite{pang2023white} introduce a white-box attack that leverages gradients at each timestep. However, all of these attack models focus on image or text-to-image diffusion models, which cannot be directly applied to GGDMs.
Table \ref{tab:comp-literature} summarizes the main differences between our work and existing studies on inference attacks against generative diffusion models.

{\bf Inference attacks on graph data.} Recent research has uncovered several types of attacks that can infer sensitive information from the training graph data. These attacks can be categorized into four types: (1) membership inference attacks (MIAs) \cite{he2021node,he2021stealing,wu2022linkteller,duddu2020quantifying, shen2022finding, wang2023link, wang2024gcl, wang2024subgraph}, seeking to determine whether a specific graph sample is part of the training dataset; (2) attribute inference attacks (AIAs) \cite{duddu2020quantifying, gong2018attribute, zhang2021graph}, aiming to infer the sensitive attributes within the training graphs; (3) property inference attacks (PIAs) \cite{wang2022group, zhang2021leakage, zhang2021inference, suri2021formalizing}, trying to infer the sensitive properties of the training graphs; and (4) graph reconstruction attacks \cite{zhang2021graphmi, shen2022finding, zhou2023strengthening}, attempting to reconstruct the training graphs. However, most of these attacks rely on the model’s final output (probability vectors) for node/graph classification \cite{he2021node, he2021stealing, wang2024subgraph, gong2018attribute, wang2022group, zhang2021leakage, suri2021formalizing}, node/graph embeddings \cite{zhang2021graph, wang2024gcl, duddu2020quantifying, shen2022finding, wang2024gcl, wang2023link, duddu2020quantifying, zhang2021inference, zhou2023strengthening}, or some other features like model gradients \cite{wu2022linkteller,zhang2021graphmi}. Consequently, these approaches are not well-suited for GGDMs due to the unique forward and reverse processes in diffusion models as well as the different types of outputs they produce.

\vspace{-0.05in}
\section{Generative Diffusion Models on Graphs}
\vspace{-0.05in}
Graph generation models aim to generate new graph samples resembling a given dataset. 
Among these, diffusion-based models have become increasingly popular. They gradually introduce noise into data until it conforms to a prior distribution \cite{niu2020permutation, huang2022graphgdp, chen2022nvdiff, jo2022score, vignac2022digress, haefeli2022diffusion, luo2022fast}. 

Generally, existing generative diffusion models on graphs include two processes: (1) the \textit{forward process}, which progressively degrades the original data into Gaussian noise \cite{ho2020denoising}, and (2) the \textit{reverse process}, which gradually denoises the noisy data back to its original structure using transition kernels. Based on how the forward and reverse processes are designed, existing generative diffusion models on graphs can be broadly categorized into three classes: {\em Score-Based Generative Models} (SGM) \cite{niu2020permutation, chen2022nvdiff}, {\em Denoising Diffusion Probabilistic Models} (DDPM) \cite{vignac2022digress, haefeli2022diffusion},
and {\em Stochastic Differential Equations} (SDE) \cite{huang2022graphgdp, jo2022score, luo2022fast}. Next, we briefly describe the forward and reverse processes for the three types of graph generative diffusion models. Note that ``synthetic graphs'' and ``generated graphs'' are used interchangeably throughout this paper to refer to the graphs produced by the diffusion models.

\begin{figure*}[t!]
    \centering
    \includegraphics[width=0.96\textwidth]{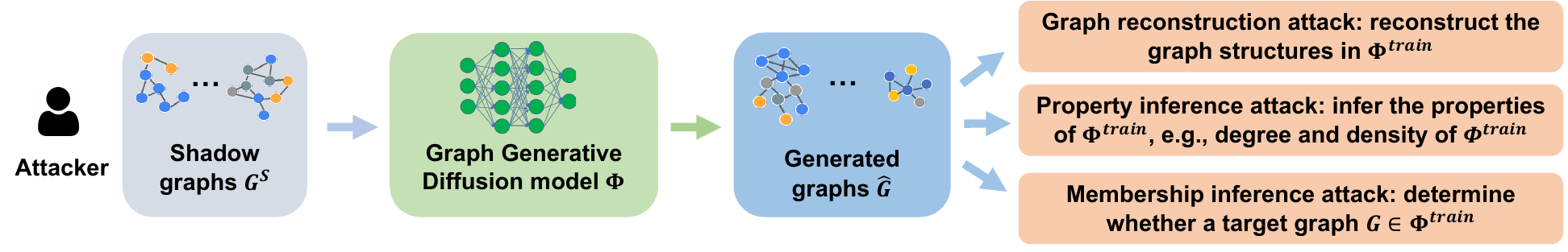}
    \vspace{-0.1in}
    \caption{\label{fig:overview} Inference attacks against graph generative diffusion models. The attacker inputs a set of shadow graphs into the target graph generative diffusion model $\TargetM$ or directly executes $\TargetM$ through an API or online marketplaces to obtain a large number of synthetic graphs $\SyntheticG$, aiming to infer the sensitive information about the training data of $\TargetM$. In this paper, we investigate three types of inference attacks: (1) reconstructing the graph structures in \Ttrain; (2) inferring the properties of \Ttrain, such as the graph density of \Ttrain; and (3) determining the membership of a given target graph.} 
     \vspace{-0.1in}
\end{figure*}
{\bf SGM-based graph generation}. The forward process injects Gaussian noise of varying intensity as the perturbation into the original graph. A noise-conditional score network is trained to represent the gradient of the conditional probability density function of the data under varying noise levels. Specifically, given a probability density function $p(x)$ and the score function $\nabla_xlog p(x)$, SGM aims to estimate the data score function in the forward process. The training objective of the score network is given by:
\vspace{-0.1in}
\begin{equation}
\label{eqn:obj-sgm}
\begin{aligned}
\small
\mathcal{L}(\theta)= \mathop{min}\limits_{\theta}
&\mathbb{E}_{t\sim\mathcal{U}(1,T),x_0\sim p(x_0),\epsilon\sim\mathcal{N}(0,\mathbf{I})}
\\
&\left[\lambda(t)\left\Vert\nabla_{{x}_t}\log p_{0t}({x}_t|x_0)-\sigma_t s_{\theta}(x_t,t)\right\Vert\right]^2,
\end{aligned}
\end{equation} 
where $\mathbb{E}$ is the expectation, $\mathcal{U}(1,T)$ is a uniform distribution over the time set $\{1, 2, ...,T\}$, $\epsilon$ is the noise vector, and $\lambda(t)$ is a positive weighting function. The variable $x_0$ refers to the original data before noise is added, with its probability density function denoted as $p(x_0)$. $p_{0t}({x}_t|x_0)$ is the score function of $x_t$. $\sigma_t$ is the Gaussian noise at $t$. The score network $s_{\theta}$ with parameter $\theta$, predicts the noise $\sigma_t$ based on $x_t$ and $t$, where $x_t$ is the noisy version of $x_0$ after adding noise at $t$.

In the reverse process, after obtaining the trained conditional score model, synthetic graphs are generated using noise-conditional score networks, such as the Score Matching with Langevin Dynamics model \cite{song2019generative} that leverages the learned score models to reconstruct graph data from noise.

\begin{table}[t!]
\centering
\setlength\tabcolsep{5pt}
\scalebox{0.92}
{\begin{tabular}{c|c}
\toprule
Symbol & Meaning \\\hline
$\TargetG$/$\ShadowG$ & Target/shadow graph\\\hline
$\TargetM$/$\ShadowM$& Target /shadow model\\\hline 
$\SyntheticG$ & Generated/synthetic graphs from $\TargetM$\\\hline
\Ttrain, \Ttest & Training and testing datasets of target model\\\hline 
\Atrain & Training set of attack model  \\\hline 
\bottomrule
\end{tabular}}
\vspace{-0.1in}
\caption{\label{tab:notation} Notations}
\vspace{-0.2in}
\end{table}
{\bf DDPM-based graph generation}. In the forward process, the original data undergoes perturbation with Gaussian noise using a fixed number. Specifically, given a probability density function $x \sim p(x)$, the forward process generates the noisy $\hat{x}$ with Markov transition kernels \cite{austin2021structured, ho2020denoising} as $p(x_t|x_{t-1})=\mathcal{N}(x_t|\sqrt{1-\beta_{t}}x_{t-1}, \beta_{t}\mathbf{I})$, where $\beta_{t}$ is a predefined variance schedule at time step $t$. 

In the reverse process, a neural network is trained to predict the noise added at each step during the forward pass, ultimately recovering the original data. 
Following the notations in Equation \ref{eqn:obj-sgm}, the optimization objective can be expressed as:
\vspace{-0.05in}
\begin{equation}
\small
\label{eqn:obj-ddpm}
\begin{aligned}
\mathcal{L}(\theta)=\mathop{min}\limits_{\theta}\mathbb{E}_{t\sim\mathcal{U}(1,T),x_0\sim p(x_0),\epsilon\sim\mathcal{N}(0,\mathbf{I})}\left[\lambda(t)\|\epsilon -\epsilon_{\theta}(x_t,t)\|^2\right],
\end{aligned}
\end{equation} 
where $\epsilon_{\theta}$ is a deep neural network with parameter $\theta$ that predicts the noise vector $\epsilon$ given $x_t$ and $t$.

{\bf SDE-based models}. The forward process employs a forward SDE \cite{li2018learning} to describe the evolution of a state variable over time to generate the noisy graph. It perturbs data to noise with SDE as $d_x=f(x,t)d_t+g(t)d_w$, where $f(x,t)$ and $g(t)$ are diffusion and drift functions of the SDE, and $w$ is a standard Wiener process.

In the reverse process, a reverse SDE is utilized to gradually convert noise to data. This is achieved by estimating the score functions of the noisy data distributions. Using similar notations as in Equation \ref{eqn:obj-sgm}, the objective for estimating the score function can be formulated as:
\vspace{-0.05in}
\begin{equation}
\label{eqn:obj-sde}
\begin{aligned}
\small
\mathcal{L}(\theta)= 
\mathop{min}\limits_{\theta}
&\mathbb{E}_{t\sim\mathcal{U}(1,T),x_0\sim p(x_0),\epsilon\sim\mathcal{N}(0,\mathbf{I})}
\\&\left[\lambda(t)\left\Vert s_{\theta}(x_t,t)-\nabla_{{x}_t}\log p_{0t}({x}_t|x_0)\right\Vert\right]^2,
\end{aligned}
\end{equation} 
Once the score function at each time step is obtained, the synthetic graphs can be generated with various numerical techniques, such as annealed Langevin dynamics, numerical SDE/ODE solvers, and predictor-corrector methods \cite{yang2023diffusion}.
\vspace{-0.12in}
\section{Motivation and Threat Model}
\vspace{-0.05in}
In this section, we start by explaining our motivation, and then define the scope and objectives of our problem. 
Table \ref{tab:notation} lists the common notations used in the paper. 

\vspace{-0.1in}
\subsection{Motivations}
\vspace{-0.05in}
Machine learning (ML) models have unlocked a variety of applications, such as data analytics, autonomous systems, and security diagnostics. 
However, developing robust models often requires substantial training datasets, extensive computational resources, and significant financial investment, which can be prohibitive for small businesses, developers, and researchers with limited budgets. Consequently, online marketplaces for ML models, such as Amazon Web Services \cite{aws}, Google AI Hub \cite{googlehub}, Modzy \cite{modzy}, Microsoft Azure Cognitive Services \cite{azure}, and IBM Watson \cite{ibm}, have emerged, facilitating model exchange, customization, and access to various machine-learning-as-a-service (MLaaS) APIs. While beneficial, these platforms raise concerns about the potential exposure of sensitive or proprietary information from the training data. 
Therefore, in this paper, we investigate the privacy vulnerability of increasingly popular GGDMs.

\vspace{-0.1in}
\subsection{Threat Model}
\vspace{-0.05in}
ML models are susceptible to privacy attacks \cite{hu2021membership,rigaki2020survey}. These attacks can be categorized into two groups based on the adversary's target \cite{rigaki2020survey}:
(1) \textit{Privacy attacks on training data}: the adversary aims to infer sensitive information about the training data.
(2) \textit{Privacy attacks on ML models}: the adversary considers the ML models themselves as sensitive, like valuable company assets, and attempts to uncover information regarding the model's architecture and parameters.

In this paper, we primarily focus on privacy attacks on training data.
Our threat model considers an adversary that interacts with a graph generative diffusion model $\TargetM$ to extract the information of the model’s training set \Ttrain. An overview of the attacks is shown in Figure \ref{fig:overview}.

\noindent{\bf Adversary’s Background Knowledge.} {We consider the adversary knowledge $\ExternalK$ along two dimensions: }
\vspace{-0.05in}
\begin{ditemize}
\item {{\em Shadow graph} $\ShadowG$ (optional): The adversary possesses one or more shadow graphs, $\ShadowG$, each with its own structure and node features. $\ShadowG$ may originate from a different domain than the model’s training set, \Ttrain, and thus exhibit distinct data distributions. In real-world scenarios, the adversary may have knowledge of a partial graph, which is a subset of the training graphs \cite{he2021stealing, wang2022group, duddu2020quantifying}. We treat this partial graph as a specific instance of a shadow graph. 
}

\item {{\em Target model} $\TargetM$: The adversary may have either white-box or black-box access to the target model. In the white-box setting, the attacker can access $\TargetM$'s internal components, such as parameters, gradients, and loss values \cite{carlini2023extracting, pang2023white, duddu2020quantifying}. In contrast, the black-box setting assumes that the adversary can only interact with $\TargetM$ through its outputs. In this paper, we consider a black-box setting, reflecting real-world scenarios such as public or commercial APIs in MLaaS platforms \cite{shokri2017membership, salem2018ml, zhang2021inference, he2021stealing}. Specifically, the adversary can submit their own graphs or random graphs to $\TargetM$ via an API and receive a corresponding set of generated graphs $\SyntheticG$ as output. This black-box setting is considered the most challenging setting for the adversary \cite{shokri2017membership, he2021stealing, salem2018ml}.
}
\end{ditemize}

\vspace{-0.05in}
\noindent{\bf Attack Goal.} We consider three types of inference attacks:
\vspace{-0.1in}
\begin{ditemize}
\item {{\em Graph reconstruction}: The attacker aims to reconstruct the graph structures within target model's training set \Ttrain. }  
\item {{\em Property inference}: The attacker attempts to infer predefined properties or aggregate characteristics, $\Property$, of an individual record or a group within the training set \Ttrain. }  
\item {{\em Membership inference}: The attacker seeks to determine whether a given graph \TargetG\ is in the training set \Ttrain. }  
\end{ditemize}

\vspace{-0.2in}
\section{Inference Attacks}
\vspace{-0.1in}
In this section, we detail the proposed graph reconstruction, property inference, and membership inference attacks. 
\begin{figure}[t!]
    \centering
    \includegraphics[width=0.47\textwidth]{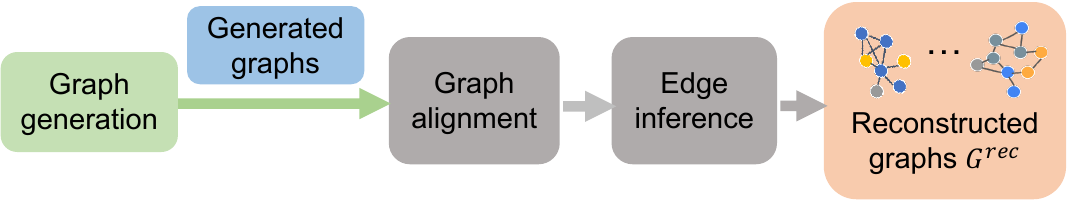}
    \vspace{-0.1in}
    \caption{\label{fig:gra overview} Overview of graph reconstruction.} 
     \vspace{-0.1in}
\end{figure}

\vspace{-0.1in}
\subsection{Graph Reconstruction Attacks}
\label{sec:GRA}
\subsubsection{Attack Overview}
Given a target GGDM $\TargetM$, obtained from an online marketplace or via an API on a MLaaS platform, the attacker's goal is to infer the graph structures in \Ttrain, which is the training set of $\TargetM$. Figure \ref{fig:gra overview} illustrates the pipeline of our graph reconstruction attack, and the corresponding pseudo-code can be found in Appendix \ref{appx:alg}.
The attacker first generates a set of synthetic graphs $\SyntheticG$ and then reconstructs the graphs by aligning each graph in $\SyntheticG$ with its closest counterpart to
identify the edges in the original training graphs. Formally, the graph reconstruction attack can be formulated as follows:
\vspace{-0.05in}
\begin{equation}
\label{eqn:gra}
f:\SyntheticG \rightarrow \Phi^{\text{train}}.
\end{equation}
Next, we detail the attack steps. 

\vspace{-0.1in}
\subsubsection{Attack Model}
Our attack model includes three steps: graph generation, graph alignment, and edge inference. 

\noindent{\bf Graph generation.} The attacker inputs a set of shadow graphs, $\ShadowG$, into $\TargetM$ or executes $\TargetM$ directly to obtain a large number of synthetic graphs $\SyntheticG$. 

\noindent {\bf Graph alignment.} With the generated graphs $\SyntheticG$, for each generated graph $g_i$ in $\SyntheticG$, we first identify the most similar graph $g_j$ to $g_i$ within $\SyntheticG$ by using graph alignment techniques \cite{yan2016short, heimann2018regal}. This is based on the assumption that if the target model has memorized a particular graph, it will likely produce multiple similar graphs to this graph. This assumption is verified in image-based diffusion models, where diffusion models memorize individual images
from their training data and emit them at generation time \cite{carlini2023extracting, van2021memorization}. 

Following the same strategy in \cite{heimann2018regal}, we use REGAL for graph alignment, which leverages representation learning to effectively map and match nodes between different graphs. Specifically, given a graph $g_i$, REGAL aligns it with other graphs in $\SyntheticG$ through the following three steps: (1) node identity extraction that extracts the structure and attribute-related information for all nodes in $g_i$ based on the degree distributions and node features; (2) similarity-based node representation by using the low-rank matrix factorization-based approach that leverages a combined structural and attribute-based similarity matrix from step (1); and (3) node representation alignment that 
greedily matches each node in $g_i$ to its top-$\alpha$ most similar nodes in other graphs in $\SyntheticG$ with k-d trees.
The difference between a pair of node representations, $(Y_i(u), Y_j(v))$, in $g_i$ and $g_j$ is calculated as: 
\vspace{-0.1in}
\begin{equation}
\label{eqn:sim}
Diff(Y_i(u),Y_j(v))=\exp^{\|Y_i(u)-Y_j(v)\|^2}.
\end{equation}

\noindent {\bf Edge inference.} After graph alignment, we get the pairwise alignments within $\SyntheticG$. Then, for each $g_i$ in $\SyntheticG$, we identify its most similar counterpart in $\SyntheticG$ by averaging all the node representation differences between the aligned graphs. Specifically, using $Diff$ from Equation \ref{eqn:sim}, we identify the most similar counterpart $\hat{g}_j$ for $g_i$ among its aligned graphs as:
\vspace{-0.1in}
\begin{equation}
\label{eqn:align}
\hat{g}_j=\mathop{min}\limits_{\hat{g}_j}\frac{1}{|V_i|}\sum_{\forall u \in \{V_i\}} Diff(Y_i(u), Y_j(\hat{u})),
\end{equation}
where $u$ is a node in the node set of $g_i$, denoted as $\{V_i\}$, $\hat{u}$ is the aligned node of $u$ in $g_j$. 

After obtaining the most similar counterpart $\hat{g}_j$ for each $g_i$ in $\SyntheticG$, and the average node representation difference between $g_i$ and $\hat{g}_j$, namely $D(g_i, \hat{g}_j)=\frac{1}{|V_i|}\sum_{\forall u \in \{V_i\}} Diff(Y_i(u), Y_j(\hat{u}))$, we pick the graph pairs with top-k\% smallest differences to determine the graph structure in \Ttrain. Empirically, we set $k$ to 10\%. Subsequently, we reconstruct the structure of a training graph from these selected aligned graph pairs. For each selected graph pair $(g_i, \hat{g}_j)$, we perform the intersection operation on the edge sets of $g_i$ and $\hat{g}_j$, denoted as ${E_i}$ and ${\hat{E}_j}$, to reconstruct the graph $g_i^{rec}$ as:
\vspace{-0.05in}
\begin{equation}
\label{eqn:edge infer}
\{V_{i}^{rec}\}=\{V_{i}\},\  
\{E_{i}^{rec}\}=\{E_{i}\} \cap \{\hat{E}_{j}\},
\end{equation}
where $\{V_{i}^{rec}\}$ and $\{E_{i}^{rec}\}$ are the node and edge sets of the reconstructed graph, respectively. We also experimented with using the union operation on the edge sets of $g_i$ and $\hat{g}_j$, but the results showed that the intersection operation performs better than the union operation. Therefore, we use the intersection operation in our attack model. Finally, we get a set of reconstructed graphs from the selected aligned graph pairs. 

\vspace{-0.1in}
\subsection{Property Inference Attacks}
\subsubsection{Attack Overview}
\begin{figure}[t!]
    \centering
    \includegraphics[width=0.4\textwidth]{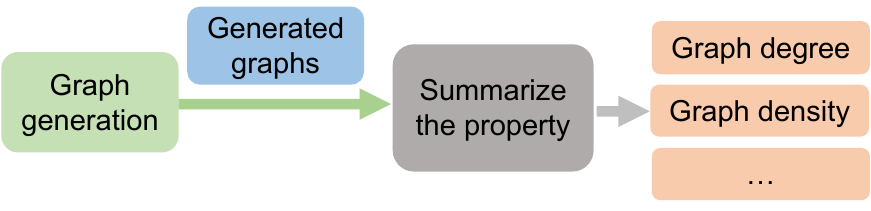}
    \vspace{-0.15in}
    \caption{\label{fig:pia overview} Overview of property inference attack.} 
     \vspace{-0.1in}
\end{figure}

Following the same attack setting, where the attacker obtains the target GGDM $\TargetM$ from an online marketplace or a MLaaS API, the attacker's goal is to infer the graph properties of \Ttrain. In this paper, we focus on two types of properties: (1) the average statistical properties of the training graphs in \Ttrain, such as graph density and average node degree; and (2) the distribution of training graphs across different property ranges in \Ttrain. Figure \ref{fig:pia overview} illustrates the pipeline of our property inference attack, and the corresponding pseudo-code can be found in Appendix \ref{appx:alg}.
The attacker first generates a set of synthetic graphs, $\SyntheticG$, and then directly calculates the property values from $\SyntheticG$.
Formally, the property inference attack can be formulated as follows:
\vspace{-0.05in}
\begin{equation}
\label{eqn:pia}
f:\SyntheticG \rightarrow \Property(\Phi^{\text{train}}),
\end{equation}
where $\Property$ represents the property, including both the average statistical properties and the distribution of training graphs across different property ranges.

\vspace{-0.05in}
\subsubsection{Attack Model}
Our attack model includes two steps: graph generation and property inference.

\noindent{\bf Graph generation.} The attacker inputs a set of shadow graphs, $\ShadowG$, into $\TargetM$ or executes $\TargetM$ directly to obtain a large number of synthetic graphs $\SyntheticG$. 

\noindent{\bf Property inference.} 
For both property types, the attacker computes the property value over all graphs in $\SyntheticG$ to approximate that of \Ttrain. The process can be formulated as:
\vspace{-0.05in}
\begin{equation}
\label{eqn:pia2}
\Property(\Phi^{\text{train}}) \leftarrow \Property(\SyntheticG).
\end{equation}

\vspace{-0.1in}
\subsection{Membership Inference Attacks}
\subsubsection{Attack Overview}
\begin{figure}[t!]
    \centering
    \includegraphics[width=0.47\textwidth]{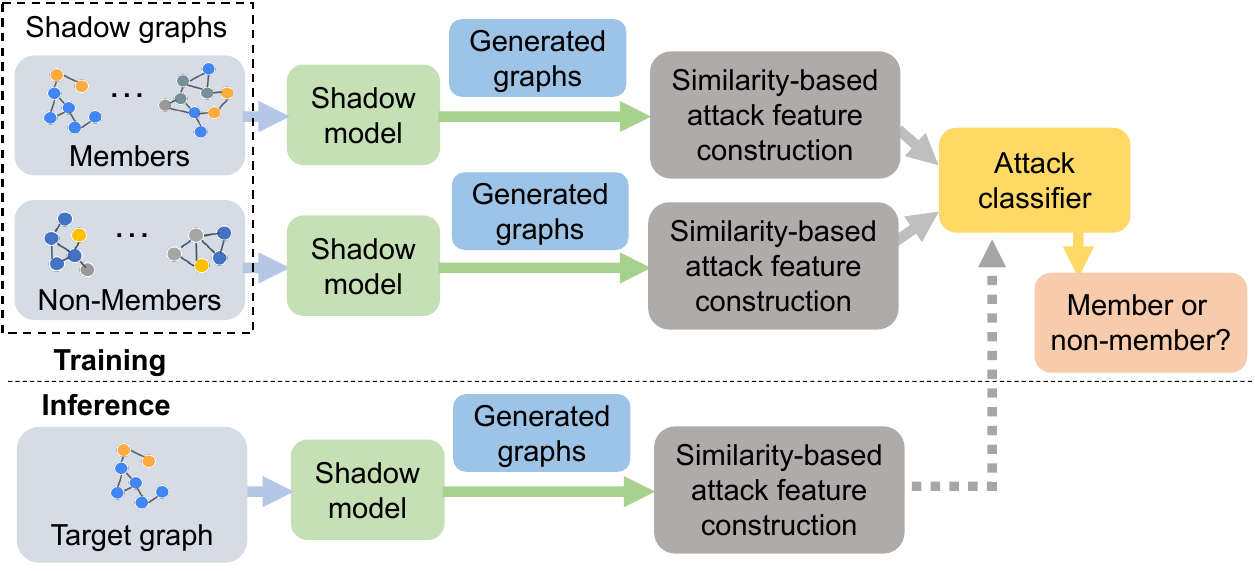}
    \vspace{-0.12in}
    \caption{\label{fig:mia overview} Overview of membership inference attack.} 
     \vspace{-0.1in}
\end{figure}

Given the GGDM $\TargetM$ obtained from an online marketplace or through an API, and a target graph $\TargetG$, the attacker's goal is to determine whether $\TargetG$ is in $\TargetM$'s training set, \Ttrain. 
In this attack, we have the following assumptions. First, the attacker has background knowledge of shadow graphs $\ShadowG$, which contain their own nodes and features. 
These shadow graphs may come from a different domain than \Ttrain. A special case occurs when $\ShadowG$ contains partial subgraphs of \Ttrain, which is plausible in real-world applications. For example, an online marketplace may release a portion of the training set. Second, following \cite{shokri2017membership, wang2022group, he2021stealing}, we assume the attacker can train a shadow model $\ShadowM$ using the same service (e.g., Google AI Hub) employed for the target model. 
Figure \ref{fig:mia overview} illustrates the pipeline of our membership inference attacks, and the corresponding pseudo-code can be found in Appendix \ref{appx:alg}.
The attacker first trains $\ShadowM$ using a subset of graphs from $\ShadowG$. 
Then, $\ShadowG$ is fed into $\ShadowM$ to generate a set of synthetic graphs $\SyntheticG$. An attack classifier is subsequently trained by analyzing the similarity between $\SyntheticG$ and $\ShadowG$ or the similarity within all the graphs in $\SyntheticG$. 
Formally, given a target graph $\TargetG$, the membership inference attack can be formulated as follows:
\vspace{-0.05in}
\begin{equation}
\label{eqn:mia}
f:\left( \TargetG, \ShadowG \right) \rightarrow y_G\in\{0,1\},
\end{equation}
where $y_G$ is the membership label of $\TargetG$, with 0 (or 1) indicating the absence (or presence) in \Ttrain. We next detail the attack.

\vspace{-0.05in}
\subsubsection{Attack Model}
Our attack model includes four steps: shadow model training and graph generation, attack feature construction, attack model training, and membership inference. 

\noindent{\bf Shadow model training and graph generation.} First, the attacker trains a shadow model $\ShadowM$ on a subset of graphs from $\ShadowG$, referred to as member shadow graphs, to mimic the behavior of the target model. The remaining graphs in $\ShadowG$ are treated as non-member shadow graphs. To balance the number of member and non-member shadow graphs, the attacker can also randomly generate non-member shadow graphs that differ from the member graphs. 
Given their large variation in node and edge sizes, the chance of overlap between member and non-member shadow graphs is negligible.
We denote the member and non-member shadow graphs in $\ShadowG$ as $\ShadowG_{mem}$ and $\ShadowG_{non-mem}$, respectively. 
Second, for each graph in $\ShadowG_i$ in $\ShadowG_{mem}$ and $\ShadowG_{non-mem}$, the attacker inputs it into $\ShadowM$ to obtain a corresponding set of synthetic graphs $\SyntheticG^S$.
We denote this as a triplet $(\ShadowG_i, \hat{G}^{S}_{i}, y)$, where $y$ is $\ShadowG_i$'s membership label.  

\noindent{\bf Attack feature construction.} After obtaining the synthetic graphs $\hat{G}^{S}_{i}$ for each graph $\ShadowG_i$ in $\ShadowG$, we construct the attack features \Atrain\ for attack training in two ways. 

{\em $\mathcal{A}^{\text{train-1}}$}: the design of this attack feature is based on the principle that 
the generated graphs $\hat{G}^{S}_{i}$ should be much similar to its corresponding shadow graph $\ShadowG_i$ when $\ShadowG_i$ is a member shadow graph in $\ShadowG$, compared to when it is a non-member. 
\underline{First}, for each shadow graph $\ShadowG_i$, the adversary measures the pairwise similarity between $\ShadowG_i$ and each generated graph $\hat{g}^S_i$ in $\hat{G}^{S}_{i}$. Specifically, we first convert both $\hat{g}^S_i$ and $\ShadowG_i$ into embedding vectors using Anonymous Walk Embeddings (AWEs) \cite{ivanov2018anonymous}, denoted as $emb({\hat{g}^S_i})$ and $emb({\ShadowG_i})$, respectively. The AWEs method represents graphs by sampling random walks and anonymizing node identities based on their first appearance indices. The frequency distribution of these anonymous walk patterns is then embedded into a lower-dimensional space using neural networks like Word2Vec. This approach captures structural similarity between graphs while being scalable and identity-independent. 
\underline{Second}, we calculate the similarity between the embedding vectors of $emb({\hat{g}^S_i})$ and $emb({\ShadowG_i})$ as $sim_k\left(emb({\hat{g}^S_i}), emb({\ShadowG_i})\right)$,
where $sim_k$ represents the $k$-th similarity function. In this
paper, $k\in\{1,2,3,4\}$ corresponds to four similarity metrics: Dot product, Cosine similarity, Euclidean distance-based difference (as defined in Equation \ref{eqn:sim}), and Jensen-Shannon Diversity (JSD). \underline{Third}, we construct the attack feature of a triplet $(\ShadowG_i, \hat{G}^{S}_{i}, y)$ by stacking the pairwise embedding similarities between $\ShadowG_i$ and $\hat{G}^{S}_{i}$, which can be written as:
\begin{equation}
\vspace{-0.1in}
\label{eqn:ft-mia1}
    A^{train}_i = 
    \begin{bmatrix}
||_{\forall k\in\{1,2,3,4\}}sim_k\left(emb({\hat{g}^S_{i,0}}), emb({\ShadowG_i})\right) \\
...\\
||_{\forall k\in\{1,2,3,4\}}sim_k\left(emb({\hat{g}^S_{i,N}}), emb({\ShadowG_i})\right)
\end{bmatrix},
\end{equation} 
where $||_{\forall k\in\{1,2,3,4\}}sim_k\left(emb(\cdot), emb(\cdot)\right)$ denotes the concatenation of similarity values computed using the four metrics between two embeddings. $\hat{g}^S{i,n}$ is the $n$-th generated graph in $\hat{G}^{S}{i}$, with $n \in {1, \dots, N}$ and $N$ being the total number of generated graphs in $\hat{G}^{S}{i}$. Therefore, for each triplet $(\ShadowG_i, \hat{G}^{S}_{i}, y)$, the dimension of its corresponding attack feature $A^{train}$ is $N \times 4$.

{\em $\mathcal{A}^{\text{train-2}}$}: the design of this attack feature is based on the principal that the generated graphs $\hat{G}^{S}_{i}$ should exhibit much higher pairwise similarity within $\hat{G}^{S}_{i}$ when $\ShadowG_i$ is a member shadow graph than when it is a non-member. 
\underline{First}, for each shadow graph $\ShadowG_i$, the adversary measures the pairwise similarity within $\hat{G}^{S}_{i}$. Similar to $\mathcal{A}^{\text{train-1}}$, we first convert each $\hat{g}^S_i$ in $\hat{G}^{S}_{i}$ into an embedding vector, $emb({\hat{g}^S_i})$, using AWEs.
\underline{Second}, we pairwisely calculate the similarity between the embedding vectors of $emb({\hat{g}^S_i})$ and $emb({\hat{g}^S_j})$ as $sim_k\left(emb({\hat{g}^S_i}), emb({\hat{g}^S_j})\right), i<j$. 
The similarity metrics $sim_k$ are the same as those used in $\mathcal{A}^{\text{train-1}}$. 
\underline{Third}, we construct the attack feature of a triplet $(\ShadowG_i, \hat{G}^{S}_{i}, y)$ by stacking the pairwise embedding similarities within $\hat{G}^{S}_{i}$:
\begin{equation}
\vspace{-0.1in}
\label{eqn:ft-mia2}
    \resizebox{.89\hsize}{!}{$A^{train}_i = 
    \begin{bmatrix}
||_{\forall k\in\{1,2,3,4\}}sim_k\left(emb({\hat{g}^S_{i,0}}), emb({\hat{g}^S_{i,1}})\right) \\
...\\
||_{\forall k\in\{1,2,3,4\}}sim_k\left(emb({\hat{g}^S_{i,N-1}}), emb({\hat{g}^S_{i,N}})\right)
\end{bmatrix}.$}
\end{equation} 
The meaning of the notations is consistent with those in Equation \ref{eqn:ft-mia1}. Therefore, for each triplet $(\ShadowG_i, \hat{G}^{S}_{i}, y)$, the dimension of its corresponding attack feature $A^{train}$ is $\frac{N*(N-1)}{2} \times 4$.

After the adversary generates the feature $A^{train}_i$ of the shadow graph $\ShadowG_i$, it associates $A^{train}_i$ with its ground-truth membership label $y$. Finally, the adversary adds the newly formed data sample ($A^{train}_i$, $y$) to \Atrain. In our empirical study, we ensure \Atrain\ is balanced, i.e., both the member and non-member classes have the same number of samples. 

\noindent{\bf Attack model training.} After \Atrain\ is generated, the adversary proceeds to train the attack classifier, such as Multi-layer Perceptron (MLP), Random Forest (RF), and Linear Regression (LR), on \Atrain. 

\noindent{\bf Membership inference.} At inference time, the adversary uses the same method used to generate the training feature $A^{train}$ to derive the feature $A^{att}$ for the target graph $\TargetG$. Specifically, the adversary inputs $\TargetG$ to the target model $\TargetM$ and obtains a set of generated graphs. Then, the adversary calculates the similarity between the generated graphs and $\TargetG$ or within the generated graphs using the same approaches and similarity functions. 
Finally, the adversary feeds $A^{att}$ into the trained attack classifier to obtain predictions, just as in the training phase. The label associated with a higher probability will be selected as the inference output.

\begin{table*}[t!]
\scalebox{0.87}{
    \centering
    \begin{tabular}{|c|c|c|c|c|c|c|c|c|c|c|c|c|c|c|c|c|}
    \hline
    \multirow{2}{*}{Dataset}&\multirow{2}{*}{Attack}&\multicolumn{5}{c|}{EDP-GNN}&\multicolumn{5}{c|}{GDSS}&\multicolumn{5}{c|}{Digress}\\\cline{3-17}
    &&P&R&F1 &$R1$&$R2$&P&R&F1 &$R1$&$R2$&P&R&F1 &$R1$&$R2$\\\hline
    \multirow{2}{*}{MUTAG}&Ours&\bf0.70&\bf0.90&\bf0.78&\bf0.21&\bf0.27&\bf0.87&\bf0.83&\bf0.85&\bf0.22&\bf0.32&\bf1&\bf0.72&\bf0.84&\bf0.11&\bf0.16\\
    &Baseline-1&0.50&0.65&0.57&0&0&0.54&0.67&0.59&0&0&0.10&0.15&0.12&0&0\\
    &Baseline-2&0.45&0.51&0.48&0&0&0.45&0.52&0.48&0&0&\bf1&0.33&0.50&0&0\\\hline
    \multirow{2}{*}{ENZYMES}&Ours&\bf0.85&\bf1&\bf0.92&\bf0.11&\bf0.36&0.83&\bf0.92&\bf0.85&\bf0.09&\bf0.28&\bf1&\bf0.79&\bf0.88&\bf0.03&\bf0.04\\
        &Baseline-1&0.54&0.73&0.62&0&0&0.75&0.75&0.70&0&0&0.01&0.19&0.01&0&0\\
&Baseline-2&0.66&0.69&0.67&0&0&\bf0.92&0.31&0.46&0&0&\bf1&0.22&0.36&0&0\\\hline
\multirow{2}{*}{Ego-small}&Ours&\bf0.84&\bf0.88&\bf0.86&\bf0.11&\bf0.24&\bf0.76&\bf0.78&\bf0.77&\bf0.07&\bf0.19&\bf1&\bf0.75&\bf0.85&\bf0.10&\bf0.22\\
&Baseline-1&0.63&0.76&0.69&0&0.03&0.60&0.72&0.65&0&0&0.09&0.43&0.15&0&0\\
&Baseline-2&0.60&0.66&0.63&0&0&0.61&0.62&0.62&0&0&\bf1&0.30&0.46&0&0\\\hline
    \multirow{2}{*}{IMDB-B}&Ours&0.85&\bf1&\bf0.92&\bf0.14&\bf0.27&\bf0.88&\bf1&\bf0.94&\bf0.15&\bf0.28&0.85&\bf0.80&\bf0.82&\bf0.09&\bf0.10\\
        &Baseline-1&0.80&0.79&0.80&0.05&0.12&0.78&0.89&0.83&0.02&0.03&0.001&0.20&0.01&0&0\\
&Baseline-2&\bf0.96&0.31&0.47&0&0&0.41&0.47&0.44&0&0&\bf1&0.21&0.34&0&0\\\hline
    \multirow{2}{*}{IMDB-M}&Ours&\bf0.92&\bf0.96&\bf0.94&\bf0.13&\bf0.19&\bf0.99&\bf1&\bf0.99&\bf0.15&\bf0.20&0.72&\bf0.89&\bf0.79&\bf0.07&\bf0.12\\
        &Baseline-1&0.73&0.83&0.78&0&0.11&0.70&0.89&0.77&0.05&0.06&0.01&0.20&0.013&0&0\\
&Baseline-2&0.61&0.78&0.68&0&0&0.54&0.58&0.56&0&0&\bf1&0.22&0.36&0&0\\\hline
\multirow{2}{*}{QM9}&Ours&\bf0.79&\bf0.84&\bf0.81&\bf0.05&\bf0.08&\bf0.80&\bf0.85&\bf0.82&\bf0.06&\bf0.11&\bf0.88&\bf0.73&\bf0.79&\bf0.05&\bf0.07\\
&Baseline-1&0.34&0.41&0.37&0&0&0.62&0.75&0.68&0&0&0.08&0.43&0.13&0&0\\
&Baseline-2&0.38&0.53&0.44&0&0&0.37&0.60&0.46&0&0&0.81&0.23&0.35&0&0\\\hline
    \end{tabular}}
    \vspace{-0.05in}
    \caption{Performance of graph reconstruction attack. "P", "R", "F1", "$R1$", and "$R2$" represent the metrics of precision, recall, F1 score, coverage ratio of the exact-matched graphs, and coverage ratio of graphs for which the attack achieves an F1 score above 0.75, respectively. For each dataset and evaluation metric under each target model, the better results are highlighted in \bf{bold}.}
    \label{tab:gra}
    \vspace{-0.1in}
\end{table*}
\vspace{-0.07in}
\section{Evaluation}
\vspace{-0.07in}
This section evaluates the effectiveness of our attacks.  

\label{sc:exp}
\vspace{-0.1in}
\subsection{Experimental Setup}
\vspace{-0.05in}
All the algorithms are implemented in Python with PyTorch and executed on NVIDIA A100-PCIE-40GB.

{\bf Datasets.} We use six real-world datasets from four domains: two molecule datasets (MUTAG, QM9), one protein dataset (ENZYMES), one citation dataset (Ego-small), and two social networks (IMDB-BINARY, IMDB-MULTI). 
These datasets each comprise a collection of graphs and serve as benchmarks for evaluating graph-based models across domains \cite{morris2020tudataset}. Statistical details are provided in Appendix \ref{appx:data}. Throughout the paper, we refer to IMDB-BINARY and IMDB-MULTI as IMDB-B and IMDB-M, respectively.

{\bf Target models.}
We employ three state-of-the-art GGDMs, namely EDP-GNN \cite{niu2020permutation}, GDSS \cite{jo2022score}, and Digress \cite{vignac2022digress}. 
\vspace{-0.07in}
\begin{itemize}[leftmargin=*]
\item {\bf EDP-GNN} is a score-based generative diffusion model (SGM) that models data distributions via score functions and represents a pioneer effort in deep graph generation.
\vspace{-0.09in}
\item {\bf GDSS} is one of the stochastic differential equations (SDE)-based models that capture the joint distribution of nodes and edges through a system of SDE. 
\vspace{-0.09in}
\item {\bf Digress} is a type of denoising diffusion probabilistic model (DDPM) that incrementally edits graphs by adding or removing edges and altering categories, subsequently reversing these changes with a graph transformer.
\end{itemize}
\vspace{-0.07in}
{\bf Implementation settings.} For the target model training, we use the default parameter settings specified in the respective papers. Both target and shadow models used early stopping: when loss fails to improve for 50 consecutive epochs. For each dataset, we randomly divided the graphs into training, validation, and testing sets with a ratio of 0.7/0.1/0.2.

\vspace{-0.05in}
\subsection{Graph Reconstruction Performance}
\vspace{-0.05in}

{\bf Evaluation metrics.} Recall that the attacker's goal is to accurately uncover the graph structures of training graphs used by the target generative models based on the generated graphs. After obtaining the reconstructed graphs $G^{rec}$, we use a graph alignment algorithm REGAL \cite{heimann2018regal} to align each reconstructed graph $g^{rec}$ in $G^{rec}$ with each training graph $g^{train}$ in \Ttrain. 
Next, we evaluate the F1 core based on all aligned pairs $(g^{rec}, g^{train})$. For each $g^{rec}$, we regard the $g^{train}$ with highest F1 score as the graph that $g^{i}$ is generated from, labeled as $g^*$. Then we calculate the attack performance as the average attack performance on all $(g^{i}, g^*)$ pairs.
\vspace{-0.07in}
\begin{itemize}[leftmargin=*]
\item {\bf Structure metrics.} We employ three edge-related metrics, namely precision (P), recall (R), and F1 score (F1), to assess the overall effectiveness of our attack in accurately recovering the exact edges and non-edges in graphs. 
\vspace{-0.09in}
\item {\bf Global Metrics.} To measure the overall graph reconstruction performance, we evaluate two coverage ratios: the proportion of training graphs exactly matched by the attack ($R1$), and the proportion with an F1 score above 0.75 ($R2$).
\end{itemize}
\vspace{-0.09in}

\begin{table*}[t!]
\scalebox{0.98}{
    \centering
    \begin{tabular}{|c|c|c|c|c|c|c|c|c|}
    \hline
    \multirow{2}{*}{Dataset}&\multicolumn{2}{c|}{EDP-GNN}&\multicolumn{2}{c|}{GDSS}&\multicolumn{2}{c|}{Digress}&\multicolumn{2}{c|}{Orig.}\\\cline{2-9}
    &$D(degree)$↓&$D(density)$↓&$D(degree)$↓&$D(density)$↓&$D(degree)$↓&$D(density)$↓&$Degree$&$Density$\\\hline
   {MUTAG}&0.009&0.001&0.123&0.026&0.139&0.013&1.093&0.070\\\hline
   {ENZYMES}&0.059&0.003&0.117&0.014&0.258&0.010&1.939&0.075\\\hline
   {Ego-small}&0.009&0.008&0.120&0.024&0.162&0.013&2.000&0.487\\\hline
    {IMDB-B}&0.109&0.008&0.178&0.010&0.190&0.029&4.850&0.219\\\hline
    {IMDB-M}&0.005&0.007&0.228&0.003&0.249&0.022&5.266&0.309\\\hline
    {QM9}&0.045&0.036&0.273&0.049&0.312&0.031&4.136&0.260\\\hline
    \end{tabular}}
    \vspace{-0.05in}
    \caption{Performance of PIA - the absolute differences in property values between the target model’s training set and the inferred values from generated graphs. A smaller difference represents a better performance. "$D(degree)$" and "$D(density)$" represent the absolute differences of average degree and density, respectively. "Orig." means the property of target model's training set \Ttrain.}
    \label{tab:pia result}
        \vspace{-0.15in}
\end{table*}

\begin{figure*}[t!]
    \centering
    \begin{tabular}{cccccc}
\multicolumn{3}{c}{\bf Average graph density}&
\multicolumn{3}{c}{\bf Average graph degree}\\
    \begin{subfigure}[b]{.14\textwidth}
      \centering
     \includegraphics[width=\textwidth]{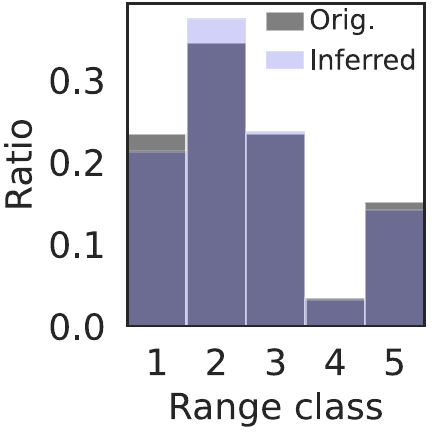}
     \vspace{-0.25in}
    \caption{\label{fig:edpgnn-den} EDP-GNN}
    \end{subfigure}
    &
      \begin{subfigure}[b]{.14\textwidth}
         \centering
    \includegraphics[width=\textwidth]{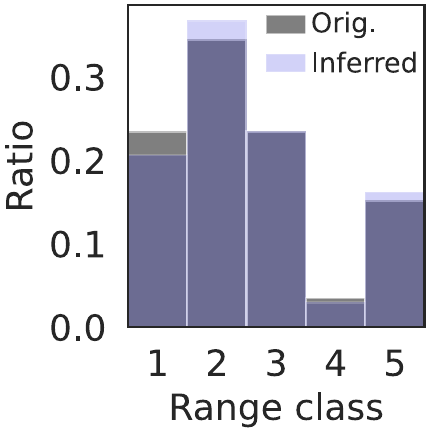}
    \vspace{-0.25in}
    \caption{\label{fig:gdss-den} GDSS}
    \end{subfigure}
    &
    \begin{subfigure}[b]{.14\textwidth}
         \centering
    \includegraphics[width=\textwidth]{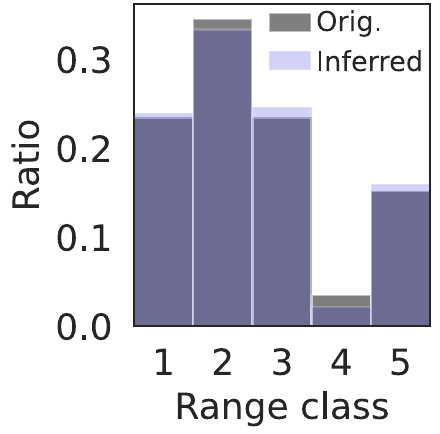}
    \vspace{-0.25in}
    \caption{\label{fig:digress-den} Digress }
    \end{subfigure}
    &
    \begin{subfigure}[b]{.14\textwidth}
      \centering
    \includegraphics[width=\textwidth]{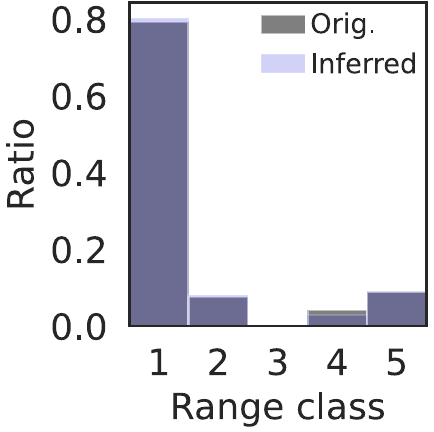}
     \vspace{-0.25in}
    \caption{\label{fig:edpgnn-dgr} EDP-GNN}
    \end{subfigure}
    &
        \begin{subfigure}[b]{.14\textwidth}
      \centering
    \includegraphics[width=\textwidth]{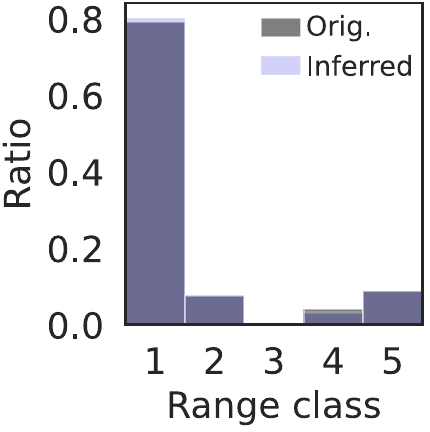}
     \vspace{-0.25in}
    \caption{\label{fig:gdss-dgr} GDSS }
    \end{subfigure}
    &
        \begin{subfigure}[b]{.14\textwidth}
      \centering
    \includegraphics[width=\textwidth]{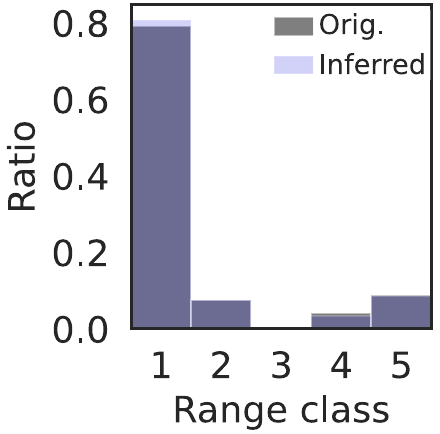}
     \vspace{-0.25in}
    \caption{\label{fig:digress-dgr} Digress}
    \end{subfigure}
    \end{tabular}
    \vspace{-0.2in}
\caption{\label{fig:pia result2} Performance of PIA -  the distribution of training graphs in different degree and density ranges. Figure (a) - (c) shows results for graph density, and Figure (d) - (f) for graph degree ($k=5$, IMDB-B dataset).} 
\vspace{-0.05in}
\end{figure*}

\noindent {\bf Attack setup.} 
For each dataset (MUTAG, Ego-small, ENZYMES, IMDB-BINARY, IMDB-MULTI, and QM9) and each target model (EDP-GNN, GDSS, and Digress), we directly run the target model $\TargetM$ and obtain 1,000 generated graphs, and then apply our GRA on these generated graphs.

\noindent {\bf Competitor.} Previous GRAs typically rely on node or graph embeddings \cite{zhang2021inference, shen2022finding}, or on the gradients of the target model (GNNs) during the training process \cite{zhang2021graphmi}. However, these approaches are not directly applicable to GGDMs, which are trained via forward and reverse processes and output a set of generated graphs. Therefore, we compare our attack with two baselines without the graph alignment step:
\vspace{-0.1in}
\begin{itemize}[leftmargin=*]
\item{\bf Baseline-1}: We treat the generated graphs $g_i \in \SyntheticG$ as replicas of \Ttrain, and evaluate attack performance as the average over all aligned pairs $(g^{i}, g^*)$, where $g^*$ in \Ttrain\ yields the highest F1 score with $g^{i}$.
\vspace{-0.1in}
\item{\bf Baseline-2}: We reconstruct a graph $g^{rec}$ by directly computing the overlapped edges between $g_i$ and $g_j$ using Equation \ref{eqn:edge infer} without graph alignment. Attack performance is evaluated over all aligned pairs $(g^{rec}, g^*)$, where $g^*$ in \Ttrain\ yields the highest F1 score with $g^{rec}$.
\end{itemize}
\vspace{-0.1in}

\noindent {\bf Experimental results.} Table \ref{tab:gra} presents the GRA results. We have the following observations. First, our attack achieves outstanding performance against all three GGDMs, with F1 scores ranging from 0.72 to 0.99, and coverage ratios of exact-matched graphs and graphs with F1 score above 0.75 reaching up to 0.21 and 0.36, respectively. Second, our attack model outperforms the baselines without the graph alignment module on structural metrics. Although Baseline-2 achieves high precision in some cases, it suffers from significantly lower recall, resulting in lower overall F1 scores. For example, the precision performance of Baseline-2 against EDP-GNN model on IMDB-B dataset is 0.96, but the recall drops to 0.31, and thus yields the F1 score of 0.47. Third, our model demonstrates substantially better global performance than both baselines, as reflected in higher coverage ratios of exact-matched graphs and graphs where the attack achieves a strong F1 score.

\vspace{-0.07in}
\subsection{Property Inference Performance}
\vspace{-0.05in}
{\bf Evaluation metrics.} To assess the attack’s effectiveness, we directly compare the property values of the target model’s training set with those inferred from the attack. The experimental results report the absolute differences between the property values of the actual training set and the inferred values from the generated graphs. 
\begin{table*}[t!]
\scalebox{0.89}{
    \centering
    \begin{tabular}{|c|c|c|c|c|c|c|c|c|c|c|}
    \hline
    \multirow{2}{*}{Dataset}&\multirow{2}{*}{Attack}&\multicolumn{3}{c|}{EDP-GNN}&\multicolumn{3}{c|}{GDSS}&\multicolumn{3}{c|}{Digress}\\\cline{3-11}
    &&$Accuracy$&$AUC$&$TPR@FPR$&$Accuracy$&$AUC$&$TPR@FPR$&$Accuracy$&$AUC$&$TPR@FPR$\\\hline
    \multirow{4}{*}{MUTAG}&Ours-1&0.650&0.763&0.225&0.821&0.811&0.467&0.845&0.881&0.327\\
    &Ours-2&\underline{0.698}&0.750&\underline{0.234}&\bf 0.887&\bf 0.936&\bf0.551&\bf 0.850&\bf 0.951&\bf 0.472\\
    &Baseline-1&\bf 0.750&\underline{0.813}&\bf 0.290&\underline{0.833}&0.889&0.519&\underline{0.849}&0.914&\underline{0.442}\\
    &Baseline-2&\bf 0.750&\bf 0.830&\bf 0.290&\underline{ 0.833}&\underline{0.917}&\underline{ 0.548}&0.833&\underline{0.944}&0.420\\\hline
    \multirow{4}{*}{ENZYMES}&Ours-1&0.813&0.831&0.427&0.825&0.868&0.467&0.817&0.845&\bf 0.344\\
    &Ours-2&\bf 0.900&\bf 0.913&\bf 0.539&\underline{0.852}&\underline{0.880}&\underline{0.562}&\underline{0.835}&\bf 0.869&\underline{0.332}\\
    &Baseline-1&0.760&0.741&0.276&0.696&0.752&0.292&0.733&0.750&0.288\\
    &Baseline-2&\underline{0.833}&\underline{0.857}&\underline{0.437}&\bf{0.855}&\bf{0.892}&\bf{0.585}&\bf{0.846}&\underline{0.868}&0.330\\\hline
    \multirow{4}{*}{Ego-small}&Ours-1&0.874&0.969&\underline{0.902}&\underline{0.917}&\underline{0.972}&0.768&\underline{0.854}&\underline{0.901}&\underline{0.453}\\
    &Ours-2&\bf0.901&\bf0.991&\bf0.913&\bf0.934&\bf0.994&\bf0.821&\bf0.892&\bf0.943&\bf0.487\\
    &Baseline-1&0.703&0.812&0.412&0.715&0.833&0.517&0.671&0.763&0.272\\
    &Baseline-2&\underline{0.882}&\underline{0.974}&0.901&0.909&0.968&\underline{0.781}&0.825&0.872&0.429\\\hline
    \multirow{4}{*}{IMDB-B}&Ours-1&\underline{0.955}&0.979&\underline{0.757}&0.908&\bf 0.999&\underline{0.957}&0.918&0.957&0.652\\
    &Ours-2&\bf0.991&\bf0.995&\bf0.878&\bf0.992&\bf0.999&\bf0.970&\underline{0.933}&\bf0.989&\bf0.727\\
    &Baseline-1&0.731&0.784&0.313&0.667&\underline{0.722}&0.288&0.640&0.703&0.292\\
    &Baseline-2&0.917&\underline{0.986}&0.733&\underline{0.986}&\bf0.999&\bf0.970&\bf0.946&\underline{0.969}&\underline{0.724}\\\hline
    \multirow{4}{*}{IMDB-M}&Ours-1&0.902&\bf0.992&\bf0.837&0.942&\underline{0.911}&0.667&0.850&0.896&0.538\\
    &Ours-2&\underline{0.912}&\underline{0.971}&\underline{0.789}&\bf0.999&\bf0.999&\bf0.985&\bf0.925&\underline{0.955}&\underline{0.633}\\
    &Baseline-1&0.739&0.781&0.308&0.724&0.762&0.302&0.645&0.711&0.298\\
    &Baseline-2&\bf 0.938&0.969&0.774&\underline{0.973}&\bf0.999&\underline{0.970}&\underline{0.892}&\bf0.961&\bf0.652\\\hline
    \multirow{4}{*}{QM9}&Ours-1&\underline{0.752}&0.816&\underline{0.412}&0.766&0.843&\underline{0.437}&0.701&0.780&\underline{0.335}\\
    &Ours-2&\bf0.781&\bf0.854&\bf0.456&\bf0.826&\underline{0.849}&\bf0.472&\bf0.772&\bf0.829&\bf0.407\\
    &Baseline-1&0.638&0.733&0.225&0.682&0.767&0.305&0.599&0.671&0.209\\
    &Baseline-2&0.750&\underline{0.821}&0.403&\underline{0.793}&\bf0.850&0.426&\underline{0.726}&\underline{0.788}&0.333\\\hline
    \end{tabular}}
    \vspace{-0.07in}
    \caption{MIA performance under setting 1 (Non-transfer). "Ours-1" and "Ours-2" denote our attacks using features $\mathcal{A}^{\text{train-1}}$ and $\mathcal{A}^{\text{train-2}}$, respectively. For each dataset and each target model, the best result is highlighted in {\bf bold}, and the second-best is \underline{underlined}.}
    \label{tab:mia}
    \vspace{-0.2in}
\end{table*}

\noindent {\bf Attack setup.} 
For each dataset and each target model, we directly run the target model $\TargetM$ and obtain 1,000 generated graphs. We calculate and summarize the property values on these generated graphs as the property of the training set of $\TargetM$. In our experiments, we consider four different types of graph properties: graph density, average node degree, average number of triangles per node, and graph arboricity (which represents the minimum number of spanning forests into which the edges of a graph can be partitioned). For each graph property, we consider two types of analysis: the average graph property across the training graphs in \Ttrain, and the distribution of \Ttrain\ across diﬀerent ranges of property values. For the second type, we uniformly bucketize the property values into $k$ distinct ranges and infer the proportion of graphs that fall into each range. We set $k=\{5,10\}$ in the experiments. 

\noindent {\bf Experimental results.} Table \ref{tab:pia result} shows the absolute differences of average graph degree and density between the target model's training set and the inferred values from generated graphs. We observe that the absolute difference in average node degree is no more than 0.312, corresponding to a difference ratio of below 7.5\% from the original value of \Ttrain. Similarly, the absolute difference in average node density is less than 0.049, with a difference ratio less than 18.8\%.
These results demonstrate that our simple yet effective attack can accurately infer the graph properties of \Ttrain.

Figure \ref{fig:pia result2} shows the distribution in different node degree and density ranges with a bucket number of $k=5$. We observe that the inferred distributions closely align with those of \Ttrain across all settings, with absolute ratio differences for each range class varying from 0.001 to 0.028 for average graph density, and 0 to 0.014 for average graph degree. 

We show the results of the properties of average number of triangles per node and graph arboricity, along with the distribution results of $k=10$ in Appendix \ref{appx:pia result2 more}. The observations are similar to those in Table \ref{tab:pia result} and Figure \ref{fig:pia result2}. 

\vspace{-0.07in}
\subsection{Membership Inference Performance}
\vspace{-0.05in}
{\bf Evaluation metrics.} For assessing MIA effectiveness, we employ three metrics: (1) {\em Attack accuracy}: the ratio of correctly predicted member/non-member graphs among all target graphs; (2) {\em Area Under the Curve (AUC)}: measured over the true positive rate (TPR) and false positive rate (FPR) at various thresholds of the attack classifier; and (3) {\em True-Positive Rate at False-Positive Rates (TPR@FPR)} \cite{carlini2022membership}: the TPR at various FPR values. We use TPR@0.1FPR in our experiments.

\noindent {\bf Attack setup.} 
For each dataset and target GGDM, we first split the original data into two halves: with 50\% used as the training set for the target generative diffusion model, representing the member graphs ($G_{mem}$), and the remaining 50\% serving as non-member graphs ($G_{non-mem}$). Next, we generate the MIA training dataset by randomly sampling 50\% of the graphs from $G_{mem}$ as members, while selecting an equal number of graphs from $G_{non-mem}$ as non-members. 
The remaining graphs from both $G_{mem}$ and $G_{non-mem}$ are used to form the MIA testing dataset. This ensures the testing set contains an equal number of member and non-member graphs, with no overlap with the training set. 
We feed each graph in the MIA training and testing set is then fed into the trained shadow model $\ShadowM$ and generate 100 graphs per input graph. These generated graphs are subsequently used to construct the attack features. We consider two distinct settings for shadow graph:
\vspace{-0.07in}
\begin{ditemize}
    \item {\bf Setting 1 (Non-transfer setting)}: Both the shadow and target graphs are sampled from the same dataset.
    \item {\bf Setting 2 (Dataset transfer setting)}: The shadow and target graphs are sampled from different datasets. Specifically, we replace the MIA testing set in the non-transfer setting with the testing set from a dataset different from that of the shadow model.
\end{ditemize}
\vspace{-0.09in}
\noindent {\bf Competitors.} We compare our method with two state-of-the-art white-box baseline attacks, both assuming the attacker has access to the target model. These baselines are regarded as strong attacks and are commonly used in prior MIA studies: one is loss-based, and the other is gradient-based.
\vspace{-0.09in}
\begin{itemize}[leftmargin=*]
\item{\bf Baseline-1: Loss-based MIA.} We adapt the white-box loss-based MIA from image-based diffusion models \cite{carlini2023extracting, matsumoto2023membership, dubinski2024towards} to our setting. First, we compute the loss values of member and non-member samples across different timesteps. Then, we train an attack classifier using the aggregated loss values from the most effective timestep range.
\vspace{-0.09in}
\item {\bf Baseline-2: Gradient-based MIA.} We adapt the white-box gradient-based MIA from image-based diffusion models \cite{pang2023white} to our setting. Specifically, we aggregate gradients from member and non-member samples over the most effective timestep range and feed them into the attack classifier.
\end{itemize}
\vspace{-0.09in}

\begin{figure*}[t!]
    \centering
    \begin{subfigure}[b]{.24\textwidth}
      \centering
     \includegraphics[width=\textwidth]{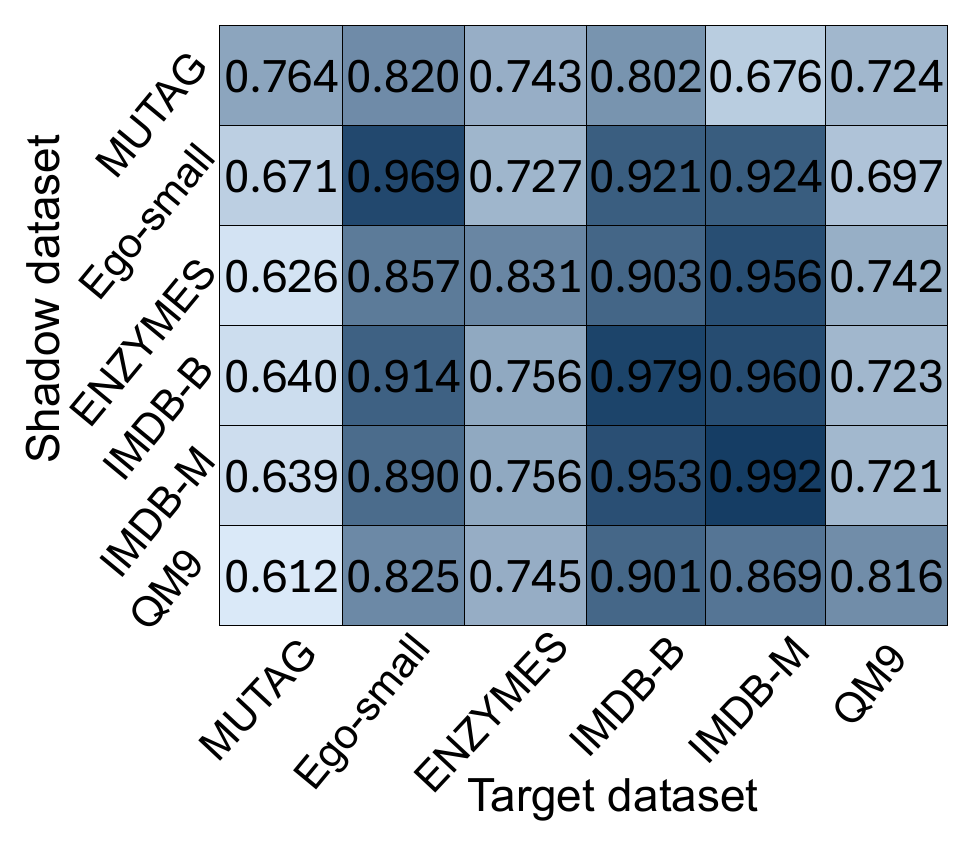}
     \vspace{-0.25in}
    \caption{\label{fig:ours1} Ours-1 }
    \end{subfigure}
      \begin{subfigure}[b]{.24\textwidth}
         \centering
    \includegraphics[width=\textwidth]{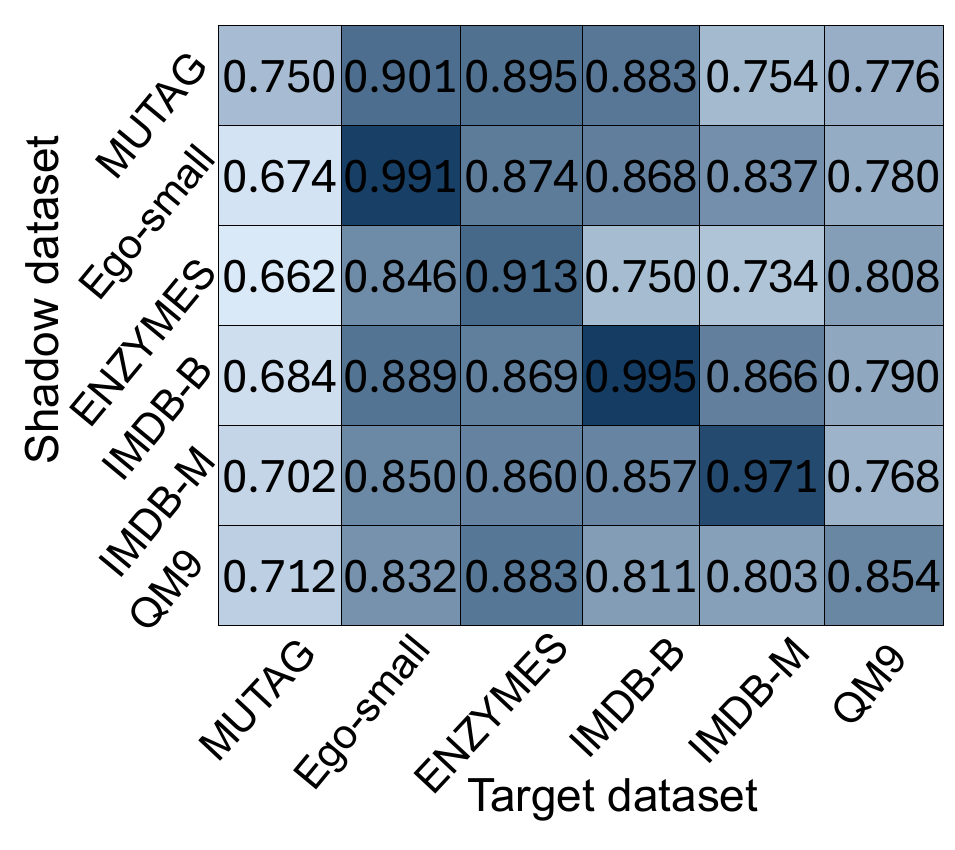}
    \vspace{-0.25in}
    \caption{\label{fig:ours2}Ours-2 }
    \end{subfigure}
    \begin{subfigure}[b]{.24\textwidth}
         \centering
    \includegraphics[width=\textwidth]{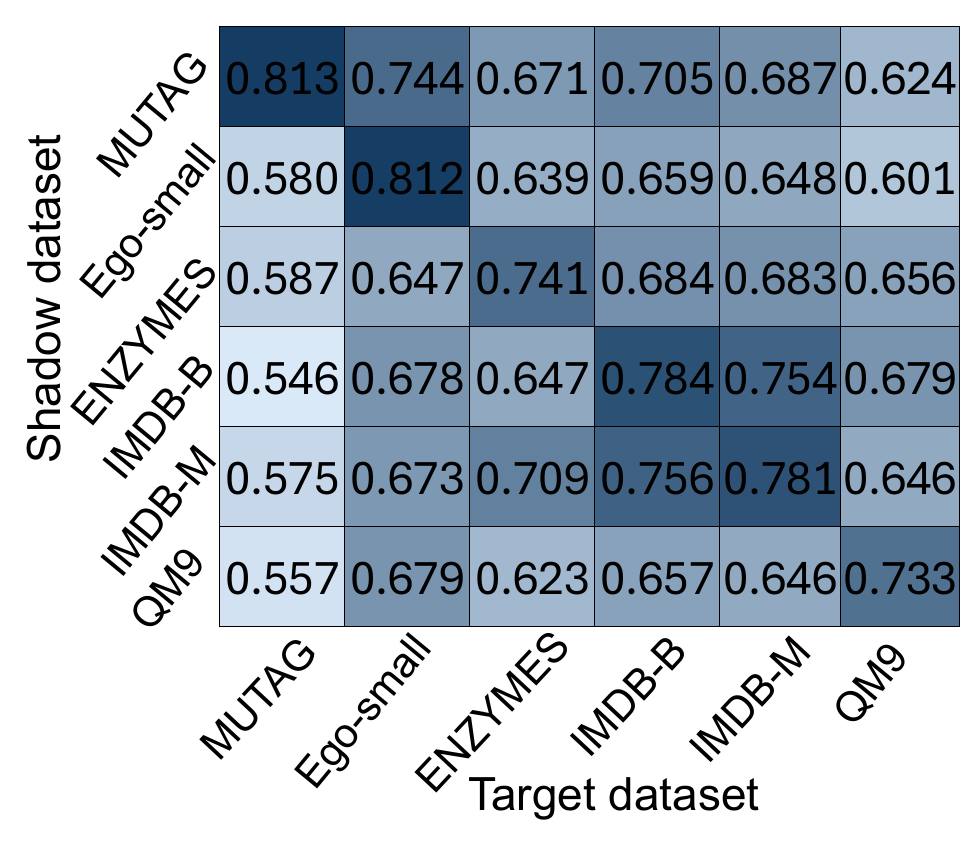}
    \vspace{-0.25in}
    \caption{\label{fig:baseline1}Baseline-1 }
    \end{subfigure}
        \begin{subfigure}[b]{.24\textwidth}
         \centering
    \includegraphics[width=\textwidth]{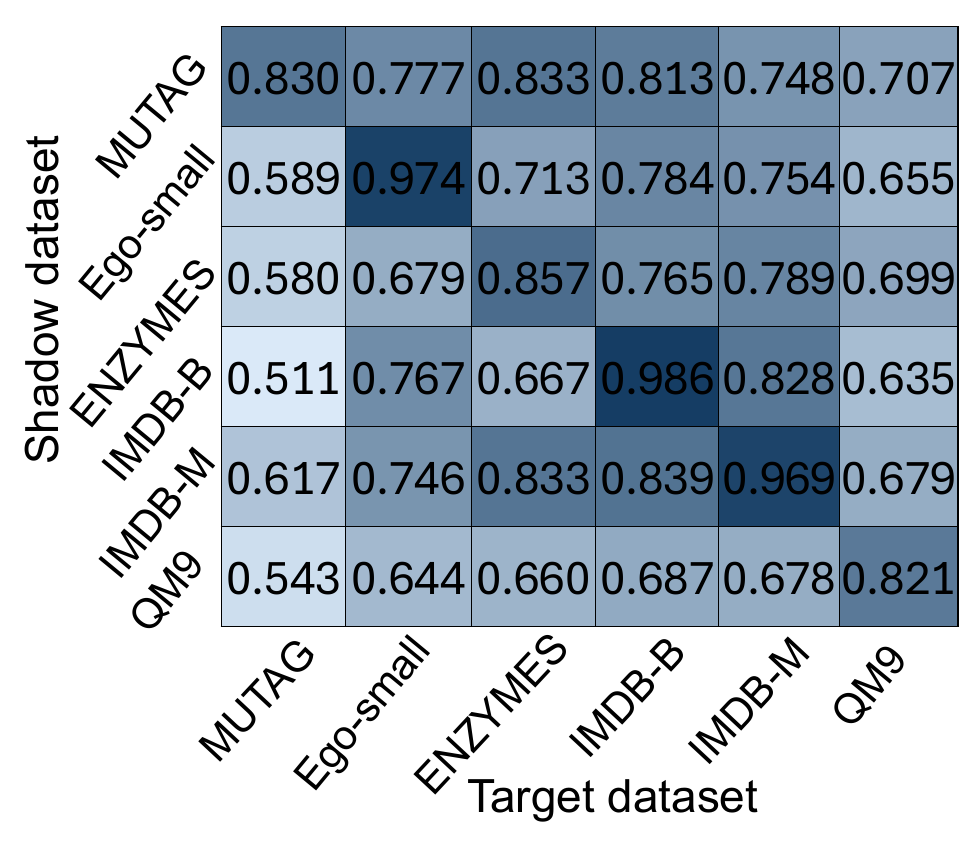}
    \vspace{-0.25in}
    \caption{\label{fig:baseline2}Baseline-2 }
    \end{subfigure}
    \vspace{-0.15in}
\caption{\label{fig:mia transfer} MIA AUC performance against EDP-GNN target model under setting 2 (dataset transfer setting).} 
\vspace{-0.05in}
\end{figure*}

\noindent {\bf Attacker performance under Setting 1 (Non-transfer setting)}. Table \ref{tab:mia} illustrates the attack performance under Setting 1. We have the following observations.
First, both our MIA models using attack feature $\mathcal{A}^{\text{train-1}}$ and $\mathcal{A}^{\text{train-2}}$ demonstrate outstanding performance across four datasets and three diffusion models, yielding attack accuracy ranging from 0.650 to 0.955 for Ours-1 and 0.698 to 0.999 for Ours-2, attack AUC ranging from 0.763 to 0.992 for Ours-1 and 0.750 to 0.999 for Ours-2, TPR@FPR ranging from 0.225 to 0.957 for Ours-1 and 0.234 to 0.985 for Ours-2. Second, Ours-2 outperforms Baselines-1 in most settings. Particularly, Ours-2's attack accuracy and AUC are respectively 0.325 and 0.277 higher than those of Baseline-1 when GDSS is the target model and IMDB-B is the target dataset. 
Third, Baseline-2 achieves performance comparable to ours in some settings, due to its use of the target model’s gradients. However, it is a white-box attack requiring access to the model’s structure and parameters, which may be difficult to obtain in real-world scenarios.
Moreover, its substantially larger feature dimension causes Baseline-2 to overfit in some settings, leading to worse performance than ours in those cases.
Additionally, Ours-2 outperforms Ours-1 in most settings, which indicates the attacker feature of $\mathcal{A}^{\text{train-2}}$ is more effective than $\mathcal{A}^{\text{train-1}}$. 

\noindent {\bf Attacker performance under Setting 2 (Data transfer setting)}. 
Figure \ref{fig:mia transfer} shows the attack results under Setting 2. The following observations can be made.
First, for each target dataset, both of our attacks achieve the highest attack accuracy when shadow and target graphs originate from the same dataset (i.e., within the same column). 
Second, even under the transfer setting, the attack remains effective, with attack AUC ranging from 0.61 to 0.96 for Ours-1 and 0.66 to 0.90 for Ours-2. This demonstrates the attack models' ability to learn knowledge from shadow data and transfer that knowledge to the target graphs. 
Specifically, for Ours-1, member graphs yield higher similarity scores between generated and input graphs than non-members; and for Ours-2, generated graphs from member graphs exhibit higher similarity than those from non-member graphs. Third, our attack models outperform both baselines in most cases, which indicates the effectiveness of our attack features. This also highlights an advantage of our model over Baseline-2, as our attacks are more applicable to real-world data transfer scenarios.

\vspace{-0.1in}
\section{Two Possible Defenses}
\vspace{-0.1in}

\begin{figure*}[t!]
    \centering
\begin{subfigure}[b]{0.27\textwidth}
      \centering
     \includegraphics[width=\textwidth]{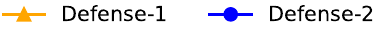}
    \end{subfigure}
    \\
    \begin{subfigure}[b]{.24\textwidth}
      \centering
     \includegraphics[width=\textwidth]{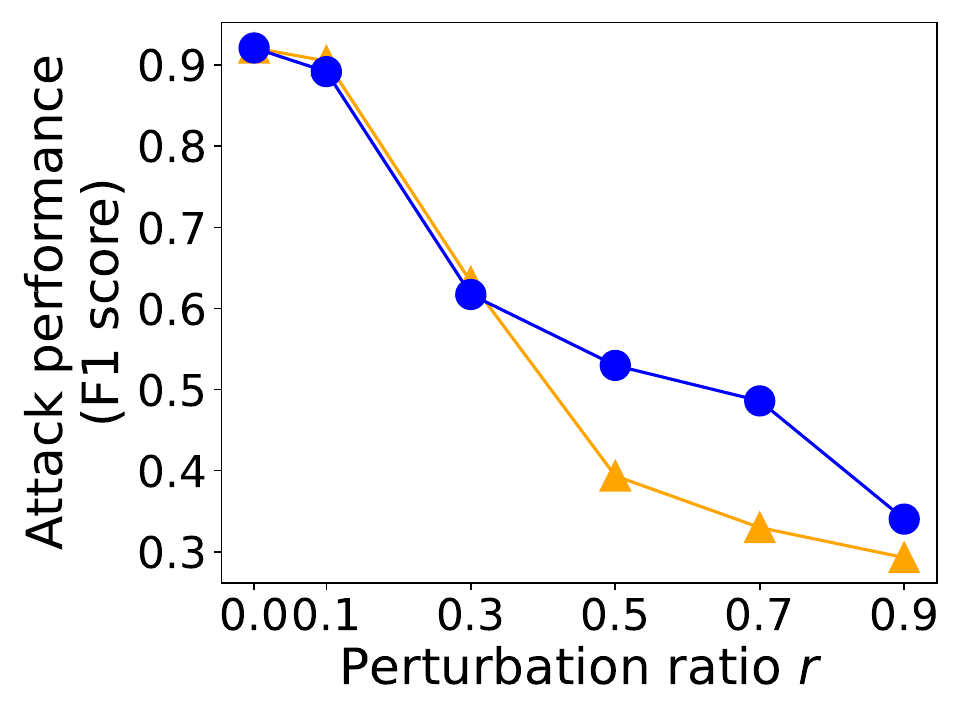}
     \vspace{-0.25in}
    \caption{\label{fig:defense-gra} GRA}
    \end{subfigure}
    \begin{subfigure}[b]{.24\textwidth}
      \centering
     \includegraphics[width=\textwidth]{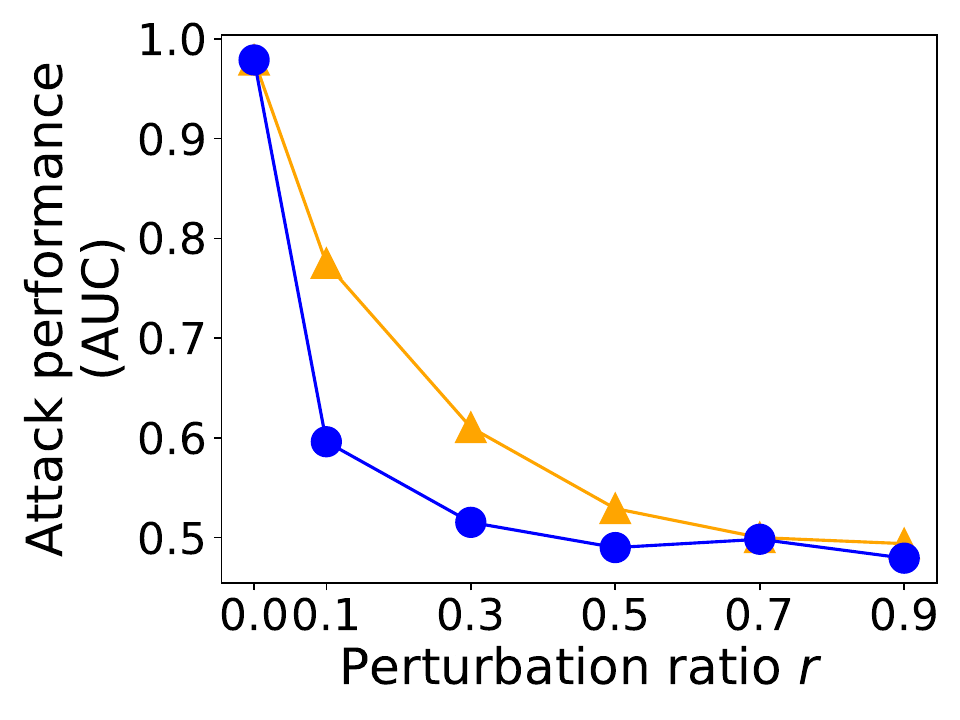}
     \vspace{-0.25in}
    \caption{\label{fig:defense-mia1} MIA-1}
    \end{subfigure}
    \begin{subfigure}[b]{.24\textwidth}
      \centering
     \includegraphics[width=\textwidth]{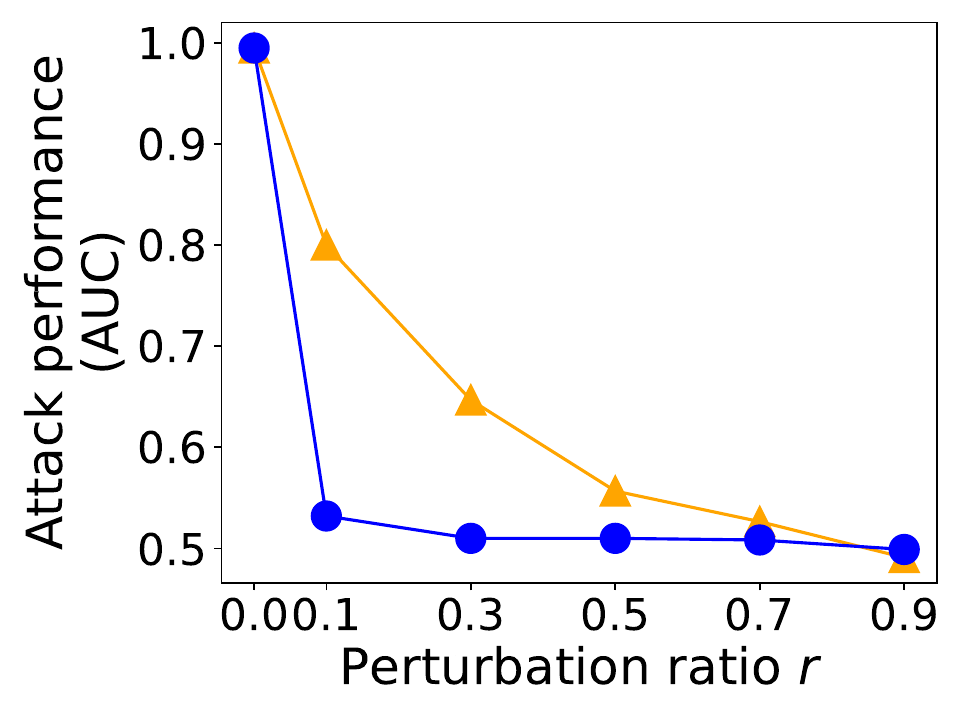}
     \vspace{-0.25in}
    \caption{\label{fig:defense-mia2} MIA-2}
    \end{subfigure}
     \begin{subfigure}[b]{.24\textwidth}
      \centering
     \includegraphics[width=\textwidth]{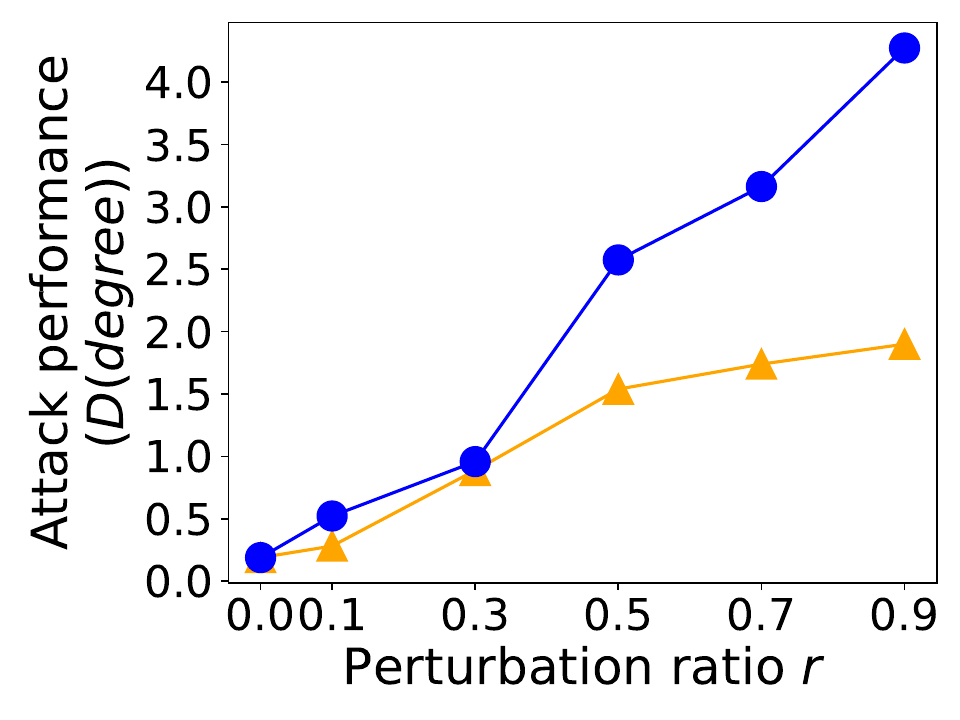}
     \vspace{-0.25in}
    \caption{\label{fig:defense-pia-deg} PIA}
    \end{subfigure}
    \vspace{-0.15in}
\caption{\label{fig:defense} Impact of perturbation ratio $r$ on defense performance (EDP-GNN model, IMDB-BINARY dataset). "MIA-1" and "MIA-2" represent MIA with attack feature of $\mathcal{A}^{\text{train-1}}$ and $\mathcal{A}^{\text{train-2}}$, respectively. $D(degree)$ denotes the difference in degree between inferred and original training graphs. For GRA, MIA-1, and MIA-2, a lower attack F1 score or AUC indicates stronger defense, while for PIA, a greater difference between the inferred and original property values represents stronger defense.} 
\vspace{-0.05in}
\end{figure*}

To enhance the privacy of GGDMs, one intuitive approach is to introduce noise into the models to confuse adversaries and provide privacy protection. Existing defense mechanisms for graph learning models mainly target GNNs \cite{niepert2016learning,hamilton2017inductive,velickovic2018graph} or random-walk-based models \cite{perozzi2014deepwalk, grover2016node2vec}. 
These strategies typically inject noise into the gradients/parameters during model training, such as equipping the model with differential privacy \cite{abadi2016deep}, or adding noise to model's output, namely the posteriors or final graph embeddings \cite{he2021stealing, shen2022finding, wang2022group}.
However, when we apply these defenses to GGDMs, the empirical evaluation (will be present in Sec. \ref{subsec:defense-results}) shows that these methods significantly degrade the quality of the generated graphs. 
To address this issue, we propose both pre-processing and post-processing defenses that perturb the structures of training graphs before model training or after graph generation, respectively. Specifically, we selectively flip the edges and non-edges that are deemed least important. This approach minimizes the quality loss of the generated graphs while providing sufficient privacy protection against inference attacks. 
We next detail the process of estimating the importance of edges and non-edges (Sec. \ref{subsec:importance}), describe our defense mechanisms (Sec. \ref{subsec:importance-defense}), and present the empirical performance (Sec. \ref{subsec:defense-results}). 

\vspace{-0.1in}
\subsection{Saliency Map-based Importance Measurement}
\label{subsec:importance}
\vspace{-0.05in}
To measure the importance of edges and non-edges, we utilize the saliency maps-based method \cite{simonyan2013deep, adebayo2018sanity, alqaraawi2020evaluating, mundhenk2019efficient}. Saliency maps are widely used in explainable machine learning and artificial intelligence, particularly in the image domain, to provide insights into the relevance of input features that contribute to a model's output \cite{mundhenk2019efficient, hsu2023explainable, szczepankiewicz2023ground}.
The most popular methods generate a gradient saliency map by back-propagating gradients from the end of the neural network and projecting it onto an image plane \cite{mundhenk2019efficient, patro2019u, simonyan2013deep}. These gradients are typically derived from a loss function, layer activation, or class activation. The absolute values of these gradients are often used to determine the importance of each feature, with a larger magnitude indicating a greater influence on the model's prediction.

In our approach, we adapt this gradient saliency map-based technique to measure the importance of edges and non-edges for node classification within each graph. This process involves three steps: forward pass through the GNNs, gradient calculation on edges and non-edges, and saliency map construction. The following sections detail each step.

{\bf Step 1: Forward pass through the GNNs.}
Give a graph $G(V, A, X, Y)$ where $V$, $A$, $X$, and $Y$ denote the nodes, adjacency matrix, node features, and node labels, respectively. We feed $G$ to a GNN for node classification. GNNs typically follow a message-passing paradigm to learn node embeddings. In this paper, we use Graph Convolutional Network (GCN) \cite{kipf2016semi}, whose layer-wise propagation is:
\vspace{-0.05in}
\begin{equation}
\label{eqn:gcn}
\small
{H}^{(l)} = \sigma\left({\hat{A}} {H}^{(l-1)} {W}^{(l)}\right),
\end{equation}
where $\hat{A}=\tilde{D}^{-1/2}(\tilde{A})\tilde{D}^{-1/2}$ denotes the normalized adjacency matrix, with $\tilde{A}=A+I$ and $I$ is the identity matrix. $\tilde{D}$ is the degree matrix of $\tilde{A}$, $W^{(l)}$ is the learnable weight matrix at $l$-th layer, and $\sigma$ denotes a non‐linear activation function such as ReLU. The ﬁnal node embeddings ${H}^{(l)}$ are passed through a softmax layer to predict node labels. 
The model is trained using cross-entropy loss.
Note that in real-world applications, some graphs may lack node features $X$ or node labels $Y$ or both. We address this by employing common strategies: (1) if $X$ is missing, we set $X$ as the identity matrix, and (2) if $Y$ is missing, we set $Y$ as the one-hot encoded node degrees.  

{\bf Step 2: Gradient calculation on edges and non-edges.} Once the GNN’s performance stabilizes after several training epochs, we backpropagate the gradient of the node classification loss to the adjacency matrix. The gradient of the loss $L$ for each entry $A_{ij}$ is computed via the chain rule as:
\vspace{-0.05in}
\begin{equation}
\label{eqn:gradient}
\small
\frac{\partial {L}}{\partial {A}_{ij}} = \sum_{k} \left( \frac{\partial {L}}{\partial {H}_k} \cdot \frac{\partial {H}_k}{\partial {\hat{A}}_{k}} \cdot \frac{\partial {\hat{A}_{k}}}{\partial {A}_{ij}} \right),
\end{equation}
where ${A}_{ij}$ is an entry in adjacency matrix ${A}$, which represents the link status between node $v_i$ and node $v_j$, $\frac{\partial {L}}{\partial {H}_k}$ denotes the gradient of loss with respect to embedding ${H}_k$ of node $v_k$, $\frac{\partial {H}_k}{\partial {\hat{A}}_{k}}$ represents the gradient of node embedding ${H}_k$ with respect to the normalized adjacency matrix, $\frac{\partial {\hat{A}_{k}}}{\partial {A}_{ij}}$ denotes the gradient of the normalized adjacency matrix with respect to ${A}_{ij}$.

{\bf Step 3: Saliency map construction.} After obtaining the gradients of all entries in ${A}$, namely the gradients of all edges and non-edges within $G$, we construct the saliency map by taking the absolute value of the gradient at each ${A}_{ij}$ as:
\vspace{-0.05in}
\begin{equation}
\label{eqn:saliency}
\text{Saliency}({A}_{ij}) = \left| \frac{\partial {L}}{\partial {A}_{ij}}  \right|.
\end{equation}
A larger gradient magnitude indicates greater importance of the corresponding edge or non-edge to model's performance.

\begin{figure*}[t!]
    \centering
\begin{subfigure}[b]{0.52\textwidth}
      \centering
     \includegraphics[width=\textwidth]{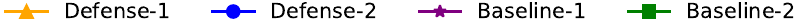}
    \end{subfigure}
    \\
    \begin{subfigure}[b]{.24\textwidth}
      \centering
     \includegraphics[width=\textwidth]{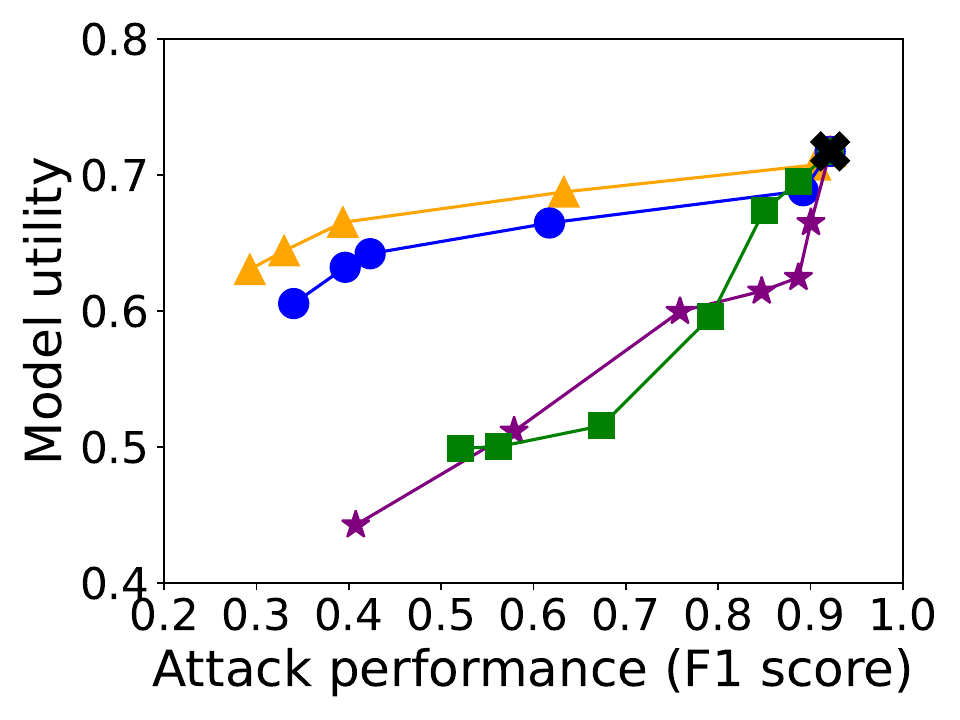}
     \vspace{-0.25in}
    \caption{\label{fig:tradeoff-gra} GRA}
    \end{subfigure}
    \begin{subfigure}[b]{.24\textwidth}
      \centering
     \includegraphics[width=\textwidth]{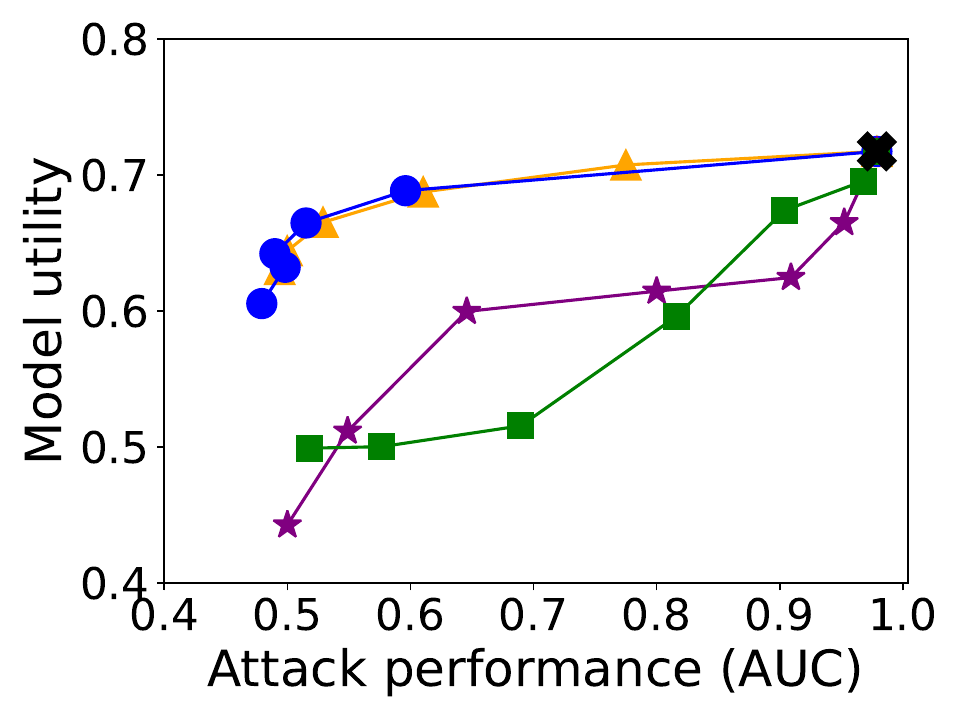}
     \vspace{-0.25in}
    \caption{\label{fig:tradeoff-mia1} MIA-1}
    \end{subfigure}
    \begin{subfigure}[b]{.24\textwidth}
      \centering
     \includegraphics[width=\textwidth]{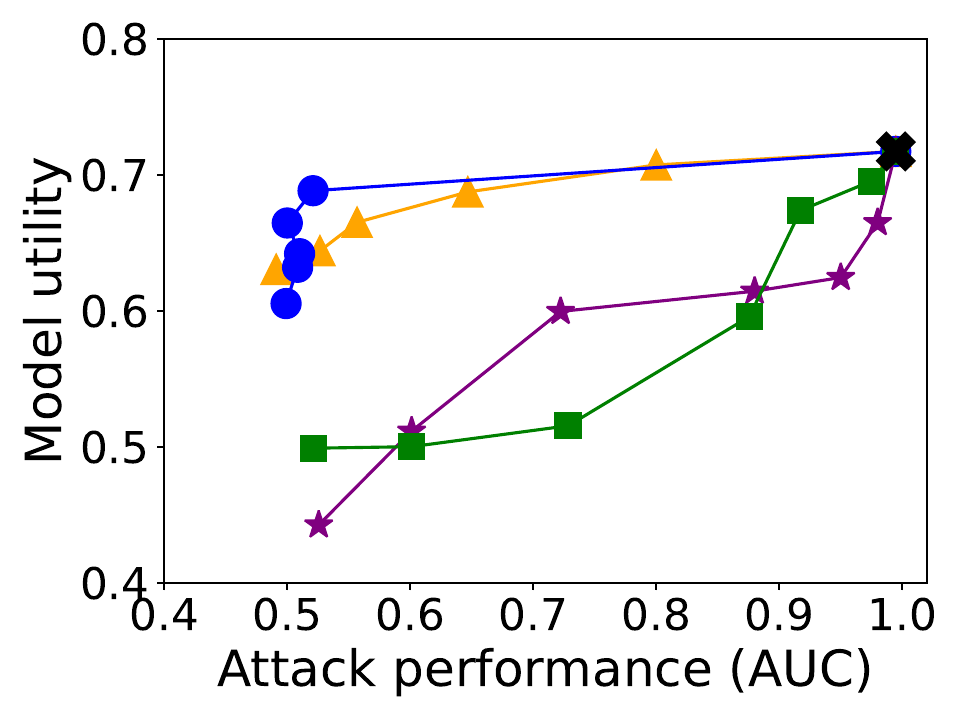}
     \vspace{-0.25in}
    \caption{\label{fig:tradeoff-mia2} MIA-2}
    \end{subfigure}
     \begin{subfigure}[b]{.24\textwidth}
      \centering
     \includegraphics[width=\textwidth]{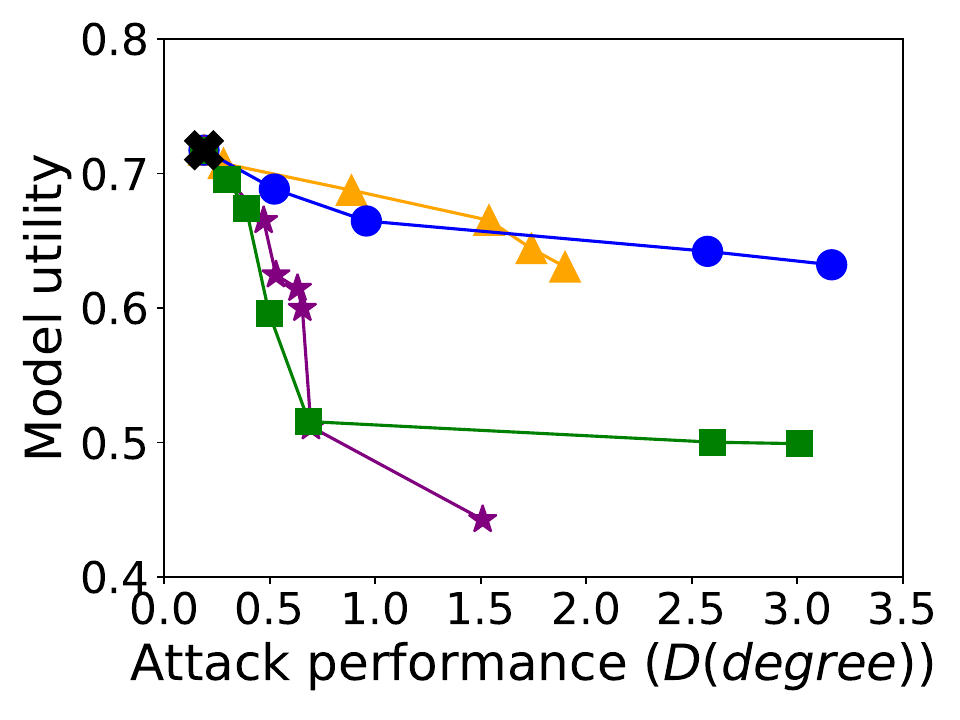}
     \vspace{-0.25in}
    \caption{\label{fig:tradeoff-pia-deg} PIA}
    \end{subfigure}
    \vspace{-0.1in}
\caption{\label{fig:tradeoff}The defense-utility curve (EDP-GNN model, IMDB-BINARY dataset). The $\x$ symbol indicates no defense. 
For GRA, MIA-1, and MIA-2, a lower attack F1 score or AUC combined with higher model utility indicates a better trade-off. For PIA, a greater difference between the inferred and original property values, along with higher model utility, signifies a better trade-off.} 
\end{figure*}
\vspace{-0.05in}
\subsection{Details of the Defenses}
\label{subsec:importance-defense}
\vspace{-0.05in}
While adding noise to training or generated graphs can reduce attacks' capabilities, random noise often causes substantial quality loss. Therefore, the primary objective of our defense is to provide effective protection against inference attacks while preserving the model utility. To achieve this objective, we propose a perturbation-based strategy that selectively flips only the least important edges or non-edges or both in the training graphs (pre-processing) or generated graphs (post-processing). Although these targeted modifications may slightly alter the utility of generated graphs, our experiments show that they effectively reduce attack performance with minimal utility loss.
Both defenses follow a three-step process. The pseudo-code of the defenses is provided in Appendix \ref{appx:alg defense}.

\vspace{0.01in}
{\bf Defense-1: Noisy training graphs (pre-processing)}. 
\vspace{-0.05in}
\begin{itemize}[leftmargin=*]
\item Step 1: For each graph $G$ in target model's training set, we construct a saliency map to measure the importance of each edge and non-edge in $G$ by using the method in Section \ref{subsec:importance}. 
\vspace{-0.22in}
\item Step 2: We rank all edges and non-edges in a single list based on their importance values in ascending order.
\vspace{-0.1in}
\item Step 3: We flip one entry in the adjacency matrix $A$ with the lowest importance. The perturbed graph is labeled as $G'$.
\end{itemize}
\vspace{-0.07in}
We then repeat Steps 1 to 3 on the newly generated graph $G'$ until the number of flipped entries reaches the pre-defined budget, $\lfloor r \times |E|\rfloor$, where $|E|$ denotes the number of edges in $G$, and $r\in(0, 1]$ is the {\em perturbation ratio}. A larger $r$ introduces more structural changes.
Finally, we train the target GGDM on the perturbed training graphs.

\vspace{0.05in}
{\bf Defense-2: Noisy generated graphs (post-processing)}.
\vspace{-0.05in}
\begin{itemize}[leftmargin=*]
\item Step 1: Similar to Defense-1, we first measure the importance of each edge and non-edge for every graph $\hat{G}$ in the generated graph set using the saliency map-based method. 
\vspace{-0.1in}
\item Step 2 and Step 3: These steps remain the same as those in Defense-1, involving ranking and flipping the least important edge or non-edge to obtain the perturbed graph $\hat{G}'$.
\end{itemize}
\vspace{-0.07in}
The process is repeated until each graph in $\hat{G}$ reaches its pre-defined flip budget, after which the perturbed graphs form the final output of the target GGDM.

\vspace{-0.15in}
\subsection{Performance of Defense}
\label{subsec:defense-results}
\vspace{-0.05in}
In this section, we provide an assessment of the performance of our defense mechanism. We use the same target models and datasets as those for the attack evaluation (Section \ref{sc:exp}).

 \noindent{\bf Setup.} We set the perturbation ratio $r=\{0, 0.1, 0.3, 0.5, 0.7, 0.9\}$. Larger values of $r$ indicate stronger privacy protection. 

\noindent{\bf Metrics.} 
We measure {\em defense effectiveness} as the attack performance after applying the defense, using the same evaluation metrics as in Section \ref{sc:exp}, with variations specific to attack types. 
The {\em utility} of the target model is evaluated as the performance of its generated graphs on a downstream task of graph classification. A higher AUC indicates better model utility.

\noindent{\bf Baselines.} For comparative analysis, we compare our defense mechanism with the following two baselines:
 \vspace{-0.06in}
 \begin{ditemize}
    \item {\bf Baseline-1: Differential privacy (DP).} We adopt DP-SGD, a state-of-the-art DP-based deep learning method \cite{abadi2016deep}, as our first baseline. We add Gaussian noise to gradients in each training iteration of target model. We set the privacy budget as $\epsilon=\{0.001, 0.01, 0.1, 1, 5, 10\}$, where a smaller $\epsilon$ indicates stronger privacy guarantees.
   \item {\bf Baseline-2: Randomly-flipped generated graphs.} 
   This baseline applies random flips where each edge and non-edge in the generated graphs is flipped with a probability $p$. We set $p=\{0.1, 0.2, 0.3, 0.4, 0.5,0.6\}$ in experiments.
\end{ditemize}
\vspace{-0.05in}
{\bf Defense effectiveness.} Figure \ref{fig:defense} presents the attack performance after deploying our defense mechanisms with various perturbation ratios when EDP-GNN is used as the target model. We have the following observations. First, our perturbation methods are highly effective against all three types of attacks, notably reducing their performance.
Second, the defense strength increases with a higher perturbation ratio $r$. Even with a modest perturbation ratio of 0.1 (perturbing only 10\% of graph edges), the attack AUC drops to around 0.6 for MIA-1 and 0.53 for MIA-2 with Defense-2. 

{\bf Defense-Utility tradeoff.} To illustrate the defense-utility trade-off, we plot the defense-utility curve of our defenses and the two baselines in Figure \ref{fig:tradeoff}. 
We establish a set of configurations by varying the perturbation ratio for our defenses, the privacy budget for Baseline-1, and the flipped probability for Baseline-2, as described in the Setup.  
For each configuration, we evaluate defense effectiveness and target model utility, with each point on the curve representing a defense-utility pair.
We observe that our defenses outperform the baselines in this trade-off. When defense performance drops below 0.6 for GRA, MIA-1, and MIA-2, which indicates ineffective attacks, our defenses maintain higher model utility compared to baselines, with degradation less than 0.05. Similarly, at the same utility level, our methods provide stronger protection across all four attacks. The superiority stems from selectively adding noise only to the least important edges and non-edges.
\vspace{-0.2in}
\section{Conclusion}
\vspace{-0.05in}
In this paper, we investigated the privacy leakage of graph generative diffusion models (GGDMs) and proposed three black-box attacks that extract sensitive information from a target model using only its generated graphs.
Specifically, we introduced (i) a graph reconstruction attack that successfully infers training graph structures of the target GGDM, (ii) a property inference attack that extracts statistical properties from the generated graphs, which closely resemble those of the training graphs, and (iii) membership inference attacks that determine whether a given graph is included in the training set of the target GGDM.
Extensive experiments demonstrate the effectiveness of these attacks against three state-of-the-art GGDMs. Furthermore, we have proposed two defense mechanisms based on perturbing training or generated graphs by flipping the least important edges and non-edges,
which have been experimentally shown to be effective. 

In future work, we will explore the vulnerability of GGDMs to other types of attacks, such as model inversion \cite{fredrikson2015model, wu2022model} and substructure inference attacks \cite{zhang2021inference, wang2024subgraph}, including strongly connected components, stars, and specialized substructures.

\vspace{-0.1in}
\section{Acknowledgments}
\vspace{-0.05in}
We thank the anonymous reviewers for their feedback. This work was supported by the Hong Kong RGC Grants RGC grants C2003-23Y and C1043-24GF. 
\newpage
\section{Ethical Considerations}
{\bf Stakeholder impact analysis.} The goal of our work is to help researchers and developers identify and reduce GGDM vulnerabilities, improving privacy posture and trustworthiness in real deployments. These vulnerabilities can affect multiple stakeholders, including GGDM deployers, data owners, and clients/users of GGDMs through the platform/API. Model deployers (e.g., cloud providers, AI marketplaces, or enterprises hosting GGDMs) may risk contract breaches, takedowns or patches, reputational harm, and disclosure obligations. Data owners may face re-identification risks and IP leakage; even without exact copies, adversaries can reverse-engineer datasets by stitching reconstructed fragments and statistics. Users may propagate sensitive priors into downstream products, and competitors can de-bias outputs to extract non-public signals. However, we have proposed two efficient defenses (Section 7) to mitigate these threats with moderate utility loss, offering practical tools for responsible GGDM deployment.

{\bf Mitigation strategies.} To minimize potential negative consequences of this research, we release defensive methods alongside the attacks so that practitioners can immediately apply appropriate mitigations. We also gate high-risk components, such as detailed attack code or sensitive hyperparameters, and provide clear guidelines for their safe and responsible use. In addition, we offer deployment recommendations, including API-level safeguards (e.g., rate limiting, authentication, anomaly detection, and detailed auditing to prevent iterative probing attacks), as well as the structure-aware output sanitization introduced in our defense section, to help strengthen real-world GGDM deployments against misuse.

{\bf Safety measures in our research process.} During research, we used public datasets and public models, no human subjects/PII, and isomorphism-invariant aggregates, avoiding reconstruction or release of any identifiable data. Our results are reported as means over multiple seeds/samples to minimize pinpointing any singular graph or any identifiable data.

{\bf Long-term commitment.} We will maintain our public repository, respond to community feedback, and update threat assessments and defense recommendations as GGDM technologies evolve. This ensures that our work continues to support safe and responsible adoption of GGDMs.
\vspace{-0.1in}
\section{Open Science}
In line with open science policies, we have made the relevant code publicly accessible to facilitate reproducibility and foster further research. As mentioned in Section 9, we release our attacks together with defenses, while withholding certain high-risk components (e.g., the AWE module in MIA) to reduce misuse. Our code is available at https://zenodo.org/records/17946102. All datasets used in this study were publicly available.
\bibliographystyle{plain}
\balance
\bibliography{bib}

\begin{algorithm}[t!]
    \caption{\label{alg:gra}
    Graph reconstruction attack}
    \KwIn{$\TargetM$ from an online marketplace or an API, shadow graphs $\ShadowG$ (optional)}
    \KwOut{The reconstructed graphs $G^{rec}$}
    Execute $\TargetM$ or feed $\ShadowG$ to $\TargetM$ to obtain the synthetic graphs $\hat{G}$\;
    Create the set of representation differences between the graphs and aligned graphs: $\D=\{\}$\;
    Create the set of reconstructed graphs: $G^{rec}=\{\}$\;
    \For{each $g_i$ in $\hat{G}$, $i\in[1,...,N]$}
        {\For{each $g_j$ in $\hat{G}$, $j\in[1,...,N], j\neq i$}
    {Do graph alignment on the graph pair $(g_i, g_j)$ with REGAL\;
    Find the most similar counterpart $\hat{g}_j$ for $g_i$ with Eq. \ref{eqn:align}
    \;   
    Add the average node representation difference between $g_i$ and $\hat{g}_j$ to $\D$, namely $\D=\D\cup D(g_i,\hat{g}_j)$\;
    }}
    Pick the graph pairs with top-k\% least differences from $\D$, label the picked graph pairs as $\D'$\;
    \For{each graph pair $(g_i, \hat{g}_j)$ in $\D'$}
    {Do edge inference with Eq. \ref{eqn:edge infer}\;
    Add the inferred graph $g_i^{rec}$ to $G^{rec}$, namely $G^{rec}=G^{rec}\cup g_i^{rec}$}
    Return $G^{rec}$
\end{algorithm}
\begin{algorithm}[t!]
    \caption{\label{alg:pia}
    Property inference attack}
    \KwIn{$\TargetM$ from an online marketplace or an API, shadow graphs $\ShadowG$ (optional), the property $\Property$ to be inferred}
    \KwOut{The inferred property of the training graphs  $\Property(\Phi^{\text{train}})$}
    Execute $\TargetM$ or feed $\ShadowG$ to $\TargetM$ to obtain the synthetic graphs $\hat{G}$\;
    Calculate the property of \Ttrain\ based on the graphs in $\SyntheticG$ with formula \ref{eqn:pia2}\;
    Return $\Property(\Phi^{\text{train}})$
\end{algorithm}
\section*{Appendix}
\appendix
\section{Pseudo code of our attacks}
\label{appx:alg}
{Algorithm \ref{alg:gra}, Algorithm \ref{alg:pia}, and Algorithm \ref{alg:mia1} show the pseudo code of our graph reconstruction, property inference, and membership inference attack, respectively.}

\vspace{-0.05in}
\section{Details of Datasets}
\label{appx:data}
\vspace{-0.05in}
We use six real-world datasets for attack evaluation in our paper, namely MUTAG, ENZYMES, Ego-small, IMDB-BINARY, IMDB-MULTI, and QM9,
which serve as benchmarks for evaluating graph-based models in various domains \cite{morris2020tudataset}. 
Table \ref{tab:data} provides details of these datasets. We obtain Ego-small and QM9 from the implementation provided by \cite{jo2022score}, while all other datasets are downloaded from the torch\_geometric package. For IMDB-BINARY and IMDB-MULTI datasets that do not have node features, we follow \cite{xie2021federated} and use one-hot degree features as the node features.

\begin{algorithm}[t!]
\small
    \caption{\label{alg:mia1}
    Membership inference attack}
    \KwIn{$\TargetM$ from an online marketplace or an API, shadow graphs $\ShadowG$, a target graph $\TargetG$}
    \KwOut{The membership label $y_G$ of $\TargetG$}
    Train a shadow model $\ShadowM$ using a subset of graphs from $\ShadowG$\;
    Attack feature matrix \Atrain$=[]$\;
    \For{each $G_i^S$ in $\ShadowG$, $i\in[1,...,N]$}{
    Feed $G_i^S$ to $\ShadowM$ to obtain a set of synthetic graphs $\hat{G}_i^S$\;
    Learn the graph embedding, $emb(G_i^S)$, of the shadow graph $G_i^S$ by using AWEs\;
    \For{each $\hat{g}_i^S$ in $\hat{G}_i^S$}
    {Learn the graph embedding, $emb(\hat{g}_i^S)$, of $\hat{g}_i^s$ by using AWEs\;}
    \If{Attack with $\mathcal{A}^{\text{train-1}}$}
    {
    \For{each embedding vector $emb(\hat{g}_i^S)$}{
    Calculate the similarity between $emb({\hat{g}^S_i})$ and $emb({\ShadowG_i})$ as $sim_k\left(emb({\hat{g}^S_i}), emb({\ShadowG_i})\right)$\;    }
    }
    \If{Attack with $\mathcal{A}^{\text{train-2}}$}{
    \For{each graph pair $(\hat{g}_i^S,\hat{g}_j^S)$ within $\hat{G}_i^S$}{
    Calculate the similarity between the embedding vectors of $emb({\hat{g}^S_i})$ and $emb({\hat{g}^S_j})$ as $sim_k\left(emb({\hat{g}^S_i}), emb({\hat{g}^S_j})\right)$\;    }}
    Construct the attack feature, $A_i^{train}$, of the triplet $(\ShadowG_i, \hat{g}_i^s, y)$ with Eq. \ref{eqn:ft-mia1} if $\mathcal{A}^{\text{train-1}}$, otherwise, use Eq. \ref{eqn:ft-mia2}, where $y$ is the known membership label of $\ShadowG_i$\;
    Append $(A_i^{train},y)$ to \Atrain\;}
    Train the attack classifier with \Atrain\;
    Feed $\TargetG$ to $\TargetM$ to obtain a set of synthetic graphs $\hat{G}$\;
    \For{each $\hat{g}$ in $\hat{G}$}
    {Learn the graph embedding, $emb(\hat{g})$, of $\hat{g}$ by using AWEs\;}
    \If{Attack with $\mathcal{A}^{\text{train-1}}$}{
    \For{each $emb(\hat{g})$ }{
    Calculate the similarity between the embedding vectors of $emb({\hat{g}})$ and $emb({\TargetG})$ as $sim_k\left(emb({\hat{g}}), emb({\TargetG})\right)$\;}
    }
    \If{Attack with $\mathcal{A}^{\text{train-2}}$}{
    \For{each graph pair $(\hat{g}_i,\hat{g}_j)$ within $\hat{G}$}{
    Calculate the similarity between the embedding vectors of $emb({\hat{g_i}})$ and $emb({\hat{g_j}})$ as $sim_k\left(emb({\hat{g_i}}), emb({\hat{g_j}})\right)$\;}}
    Construct the attack feature, $A^{att}$ of $\TargetG$ using Eq. \ref{eqn:ft-mia1} when adopting $\mathcal{A}^{\text{train-1}}$, otherwise, use Eq. \ref{eqn:ft-mia2}\;
    Feed $A^{att}$ to the trained attack classifier to obtain the predicted membership $y_G$\;
    Return $y_G$
\end{algorithm}

\section{More results on property inference attack}
\label{appx:pia result2 more}
{\bf PIA performance on the properties of average number of triangles per node and graph arboricity.} 
Table \ref{tab:pia result2} presents the absolute differences in the average number of triangles per node and the average graph arboricity between the target model's training set, \Ttrain, and the inferred values from the generated graphs. We observe that the absolute difference in the average number of triangles per node does not exceed 1.005 (less than 3.92\% of its original value in \Ttrain). Similarly, the absolute difference in average graph arboricity remains below 0.476 (less than 9.05\% of its original value in \Ttrain).
These results demonstrate that our simple yet effective attack can accurately infer the graph properties of \Ttrain.

\begin{algorithm}[h]
    \caption{\label{alg:defense1}
    \small
    Graph perturbation in the defense}
    \KwIn{The graphs to be perturbed, namely, the training graphs of GGDM in Defense-1 or the synthetic graphs in Defense-2, are denoted as $\G$. The perturbation ratio $r$. Note that we perturb only the graph structures of the graphs in $\G$, while other properties unrelated to the structure, such as node features, are kept unchanged. Therefore, we omit these properties in the following code.}
    \KwOut{The perturbed graphs $\G'$}
    \For{each graph $g$ in $\G$}
        {Calculate the perturbation budget $b=\lfloor r\times|E| \rfloor$, where $|E|$ is the number of edges in $g$\;
        $\P=\{\}$, used to label the perturbed entries in the adjacency matrix $A$ of $g$\; 
        $\G'=\{\}$\;
        \While{b>0}{
        Perform a forward pass of $g$ through GCN for 10 epochs\;
        Calculate the node classification loss $L$ and backpropagate the gradient to each entry in $A$ using Eq. \ref{eqn:gradient}\;
        Construct the saliency map $Saliency(A)$ with Eq. \ref{eqn:saliency} 
        \;
        Rank the entries in $A$ based on their importance values in $Saliency(A)$ in ascending order, labeled as $A_{ranked}$\;
        \For{$A_{ij}$ in $A_{ranked}$}
        {\If{$A_{ij} \notin \P$}
        {Flip the $A_{ij}$ in $A$\;
        $\P=\P \cup A_{ij}$\;
        Continue;
        }
        }       
        $b=b-1$\;
        }
        $\G'=\G'\cup A$\;
        }
    Return $\G'$
\end{algorithm}
\begin{table*}[t!]
\scalebox{0.89}{
    \centering
    \begin{tabular}{|c|c|c|c|c|c|c|c|c|}
    \hline
         \textbf{Dataset} &\textbf{Domain}& \textbf{\# Graphs} & \textbf{Avg. \# Nodes}  & \textbf{Avg. \# Edges} & \textbf{Avg. \# Degrees} & \textbf{\# Node features} &\textbf{\# Classes}\\\hline
         \textbf{MUTAG}&Molecules&188&17.93&19.79&2.11&7&2\\     \textbf{Ego-small}&Citation network&200&6.41&8.69&2.00&3,703&-\\ \textbf{ENZYMES}&Proteins&600&32.60&124.3&7.63&3&6\\
         \textbf{IMDB-BINARY}&Social network&1,000&19.77&96.53&9.77&135&2\\
         \textbf{IMDB-MULTI}&Social network&1,500&13.00&65.94&10.14&88&3\\
         \textbf{QM9}&Molecules&133,855&18.03&37.30&4.14&11&-\\

\hline
    \end{tabular}}
       \vspace{-0.05in}
    \caption{Statistics of datasets}
    \label{tab:data}
   \vspace{-0.1in}
\end{table*}

\begin{figure*}[t!]
    \centering
    \begin{tabular}{cccccc}
\multicolumn{3}{c}{\bf Density}&
\multicolumn{3}{c}{\bf Degree}\\
    \begin{subfigure}[b]{.14\textwidth}
      \centering
\includegraphics[width=\textwidth]{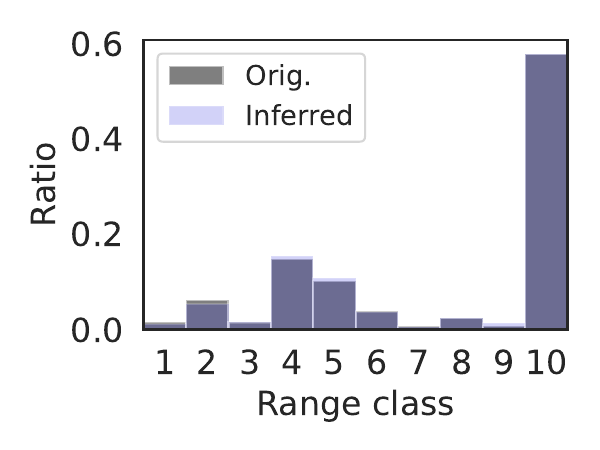}
     \vspace{-0.3in}
    \caption{\label{fig:edpgnn-den10} EDP-GNN}
    \end{subfigure}
    &
      \begin{subfigure}[b]{.14\textwidth}
         \centering
\includegraphics[width=\textwidth]{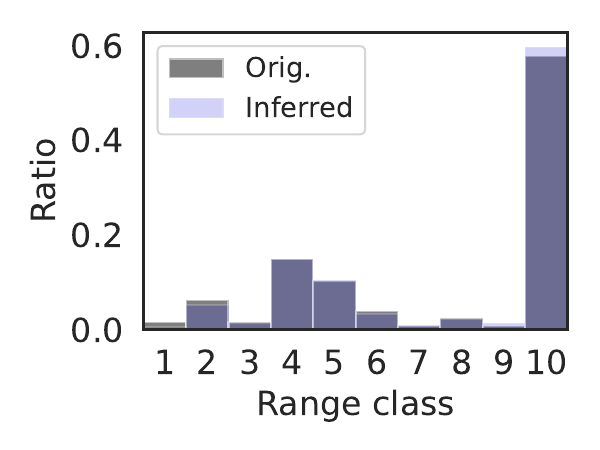}
    \vspace{-0.3in}
    \caption{\label{fig:gdss-den10} GDSS}
    \end{subfigure}
    &
    \begin{subfigure}[b]{.14\textwidth}
         \centering
\includegraphics[width=\textwidth]{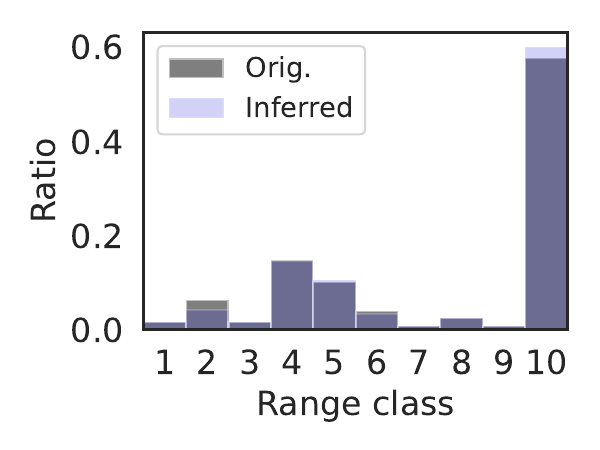}
    \vspace{-0.3in}
    \caption{\label{fig:digress-den10} Digress}
    \end{subfigure}
    &
    \begin{subfigure}[b]{.14\textwidth}
      \centering
\includegraphics[width=\textwidth]{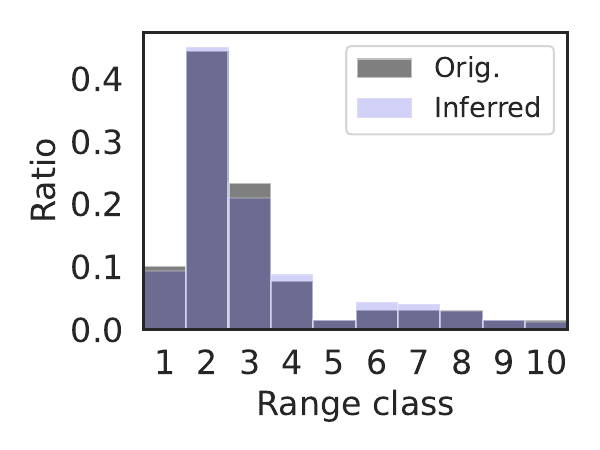}
     \vspace{-0.3in}
    \caption{\label{fig:edpgnn-dgr10} EDP-GNN}
    \end{subfigure}
    &
        \begin{subfigure}[b]{.14\textwidth}
      \centering
    \includegraphics[width=\textwidth]{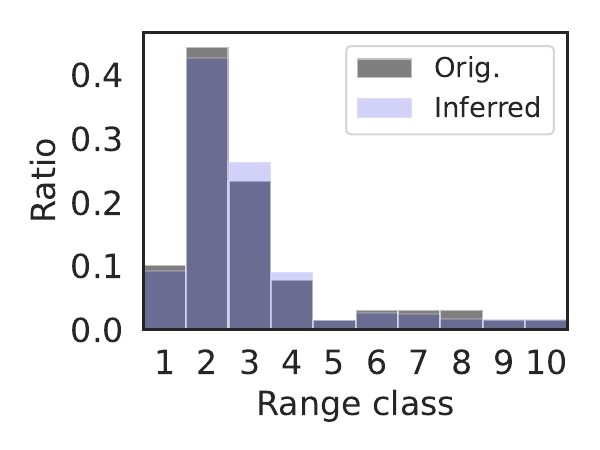}
     \vspace{-0.3in}
    \caption{\label{fig:gdss-dgr10} GDSS}
    \end{subfigure}
    &
        \begin{subfigure}[b]{.14\textwidth}
      \centering
\includegraphics[width=\textwidth]{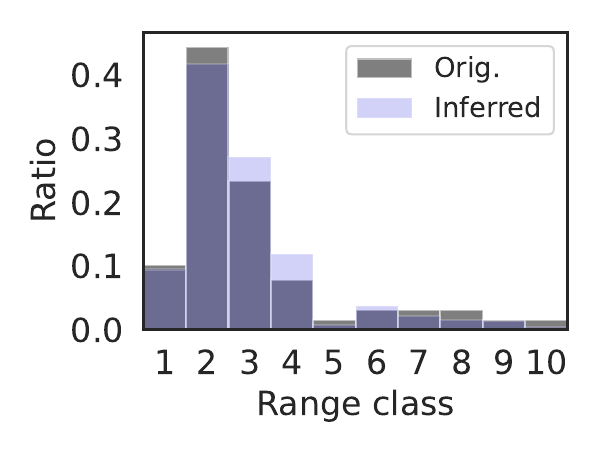}
     \vspace{-0.3in}
    \caption{\label{fig:digress-dgr10} Digress}
    \end{subfigure}
\\
\multicolumn{3}{c}{\bf \# Triangles}&
\multicolumn{3}{c}{\bf Arboricity}\\
    \begin{subfigure}[b]{.14\textwidth}
      \centering
    \includegraphics[width=\textwidth]{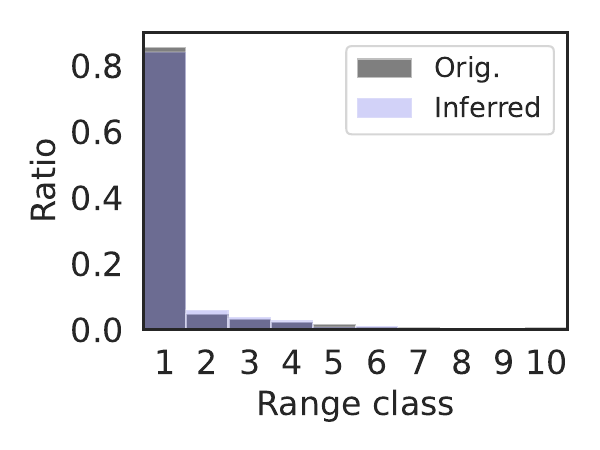}
     \vspace{-0.3in}
    \caption{\label{fig:edpgnn-triangle10} EDP-GNN}
    \end{subfigure}
    &
      \begin{subfigure}[b]{.14\textwidth}
         \centering
    \includegraphics[width=\textwidth]{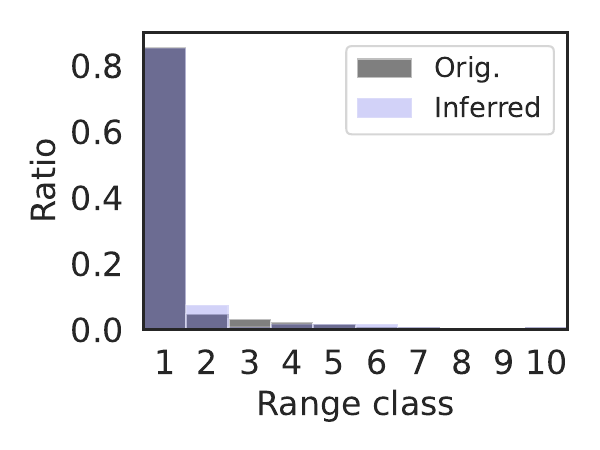}
     \vspace{-0.3in}
    \caption{\label{fig:gdss-triangle10} GDSS}
    \end{subfigure}
    &
    \begin{subfigure}[b]{.14\textwidth}
         \centering
    \includegraphics[width=\textwidth]{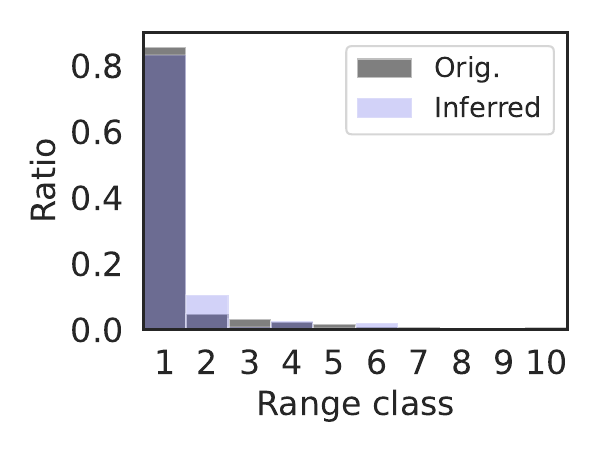}
     \vspace{-0.3in}
    \caption{\label{fig:digress-triangle10} Digress}
    \end{subfigure}
    &
    \begin{subfigure}[b]{.14\textwidth}
      \centering
    \includegraphics[width=\textwidth]{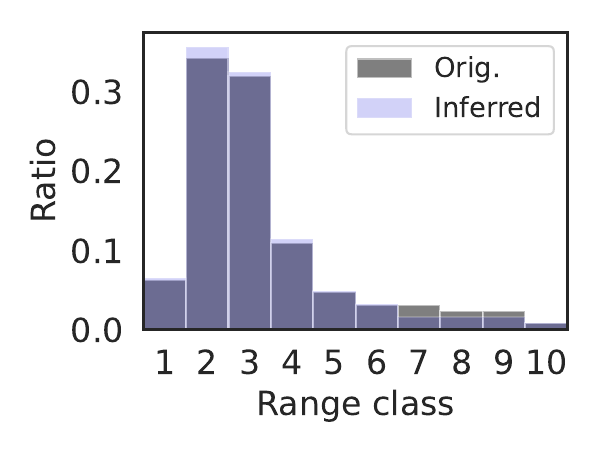}
     \vspace{-0.3in}
    \caption{\label{fig:edpgnn-arboricity10} EDP-GNN}
    \end{subfigure}
    &
        \begin{subfigure}[b]{.14\textwidth}
      \centering
    \includegraphics[width=\textwidth]{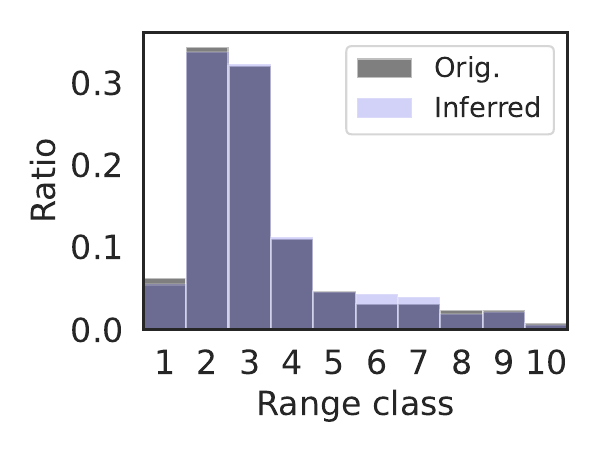}
     \vspace{-0.3in}
    \caption{\label{fig:gdss-arboricity10} GDSS}
    \end{subfigure}
    &
        \begin{subfigure}[b]{.14\textwidth}
      \centering
    \includegraphics[width=\textwidth]{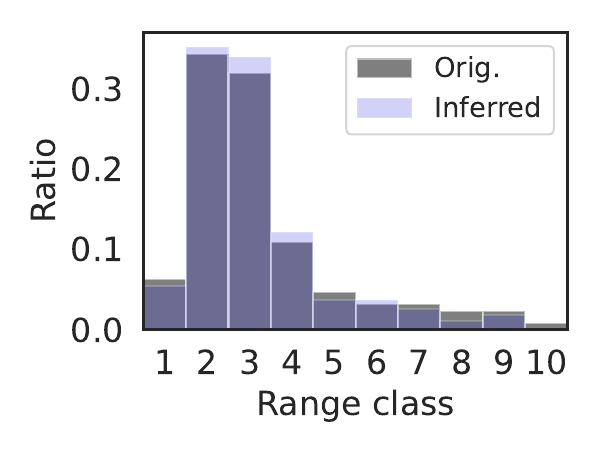}
     \vspace{-0.3in}
    \caption{\label{fig:digress-arboricity10} Digress}
    \end{subfigure}
    \end{tabular}
    \vspace{-0.1in}
\caption{\label{fig:pia result2-10} Performance of PIA - the portion of training graphs in different property ranges. Figures (a)–(c) report attack performance on graph density, Figures (d)–(f) on graph degree, Figures (g)–(i) on the average number of triangles per node, and Figures (j)–(l) on graph arboricity ($k=10$, IMDB-M dataset).} 
\end{figure*}
\begin{table*}[t!]
\scalebox{0.85}{
    \centering
    \begin{tabular}{|c|c|c|c|c|c|c|c|c|}
    \hline
    \multirow{2}{*}{Dataset}&\multicolumn{2}{c|}{EDP-GNN}&\multicolumn{2}{c|}{GDSS}&\multicolumn{2}{c|}{Digress}&\multicolumn{2}{c|}{Orig.}\\\cline{2-9}
    &$D(triangle)$↓&$D(arborivity)$↓&$D(triangle)$↓&$D(arborivity)$↓&$D(triangle)$↓&$D(arborivity)$↓&$\# Triangles$&$Arborivity$\\\hline
   {MUTAG}&0.034&0.0001&0.292&0.522&0.398&0.044&0&1\\\hline
   {ENZYMES}&0.143&0.041&0.212&0.158&0.135&0.245&1.035&1.607\\\hline
   {Ego-small}&0.203&0.037&0.257&0.174&0.286&0.226&4.605&1.175\\\hline
    {IMDB-B}&0.914&0.103&0.828&0.351&0.881&0.405&20.167&4.611\\\hline
    {IMDB-M}&0.448&0.048&0.917&0.351&1.005&0.467&25.627&5.161\\\hline
    {QM9}&0.374&0.055&0.413&0.219&0.580&0.324&11.185&2.225\\\hline
    \end{tabular}}
    \vspace{-0.1in}
    \caption{Performance of property inference attack, namely the absolute differences in property values between the target model’s training set and the inferred values from the generated graphs. A lower difference represents a better performance. "$D(triangle)$" and "$D(arborivity)$" represent the absolute difference of average number of triangles per node and graph arborivity, respectively. "Orig." means the property of target model's training set \Ttrain.}
    \label{tab:pia result2}
    \vspace{-0.05in}
\end{table*}

{\bf PIA performance on frequency ratio distribution across different property ranges.} Figure \ref{fig:pia result2-10} shows the frequency distribution in different property ranges with a bucket number of $k=10$. We observe that the inferred frequency distribution closely aligns with the original distribution of \Ttrain across all settings, with absolute ratio differences for each range class varying from 0.0003 to 0.0235 for average graph density, 0 to 0.0401 for average graph degree, 0 to 0.0589 for average number of triangles per each node, and 0.0003 to 0.0202 for average graph arboricity. 

\section{Pseudo code of our defense mechanism}
\label{appx:alg defense}
Algorithm \ref{alg:defense1} shows the pseudo code for the graph perturbation part of our two defense mechanisms.

\end{document}